\documentclass{article} 
\usepackage{iclr2027_conference,times}

\usepackage{amsmath,amsfonts,bm}

\def\eqref#1{equation~\ref{#1}}

\def\1{\bm{1}}

\DeclareMathAlphabet{\mathsfit}{\encodingdefault}{\sfdefault}{m}{sl}
\SetMathAlphabet{\mathsfit}{bold}{\encodingdefault}{\sfdefault}{bx}{n}

\usepackage{hyperref}
\usepackage{url}
\usepackage{graphicx}
\usepackage{subcaption}
\usepackage{wrapfig}
\usepackage{booktabs}
\usepackage{multirow}
\usepackage{makecell}
\usepackage[dvipsnames]{xcolor}
\usepackage{colortbl}

\title{FastBench: Can Streaming VLMs Perceive High-Dynamic Real-World Streams?}

\author{%
\normalfont\itshape Yuxuan Hu\textsuperscript{1}\thanks{Equal contribution.}, Weikang Shi\textsuperscript{1,*}, Yang Bo\textsuperscript{2}, Xudong Lu\textsuperscript{1}, Xintong Guo\textsuperscript{2}, \\
\normalfont\itshape Shuhan LI\textsuperscript{2}, Yuyang He\textsuperscript{2}, Huankang Guan\textsuperscript{2}, Peiwen Sun\textsuperscript{1}, \\
\normalfont\itshape Yunqiao Yang\textsuperscript{1}, Wenbo Li\textsuperscript{1}, Rui Liu\textsuperscript{2}, Hongsheng Li\textsuperscript{1} \\[0.5em]
\normalfont\textsuperscript{1}CUHK MMLab \\
\normalfont\textsuperscript{2}Huawei Research
}

\iclrfinalcopy 
\begin{document}

\maketitle
\lhead{Preprint} 

\begin{abstract}
Streaming Video Large Language Models (VLMs) have emerged as a promising paradigm for real-time, continuous video understanding. However, existing streaming benchmarks predominantly target low-dynamic scenarios, masking a critical challenge: under a bounded context budget, a model must trade off long temporal history, spatial resolution, and temporal granularity, and the sparse sampling rates (1--2 FPS) adopted by current systems inevitably miss fast-paced events. To bridge this gap, we introduce FastBench, a benchmark designed to evaluate streaming VLMs on high-dynamic real-world video streams. FastBench is built with a trajectory-grounded pipeline: a VLM proposes candidate QA pairs from native high-frame-rate clips, a low-FPS filter discards pseudo-dynamic questions that remain answerable at 2 FPS, and a Vision Expert Verification stage grounds answers on object trajectories extracted by SAM3 and CoTracker3 to suppress hallucinations, followed by three rounds of human inspection. The resulting 306 QA pairs span eight domains (e.g., sports, gaming, wildlife, and transportation), six capability dimensions, and forward, instant, and backward temporal scopes, each with human-annotated evidence intervals. Alongside the benchmark, we present ProactiveFrame, a training-free baseline that lets a model raise or restore the frame rate of incoming chunks through text tokens, supported by a dual-tier sliding window that buffers recent high-FPS chunks and gracefully degrades them into sparse history. Experiments show that high-dynamic perception remains far from solved: the strongest model, Gemini-3.5-Flash, reaches only 50.7\%. Denser sampling helps substantially (Qwen3-VL-8B improves from 32.9\% at 2 FPS to 44.6\% at 24 FPS) but saturates as history is compressed. ProactiveFrame improves over sparse uniform sampling (+5.4\% and +1.5\%), yet stays well below oracle-guided focusing, revealing that current VLMs struggle to decide, from the visual stream alone, when finer temporal perception is needed. FastBench establishes a rigorous testbed and offers actionable insights for high-dynamic streaming video understanding.
Our project is open source and available at \url{https://github.com/Ashone3/FastBench}.
\end{abstract}

\section{Introduction}

The remarkable success of Large Language Models (LLMs) and Vision-Language Models (VLMs) has catalyzed significant progress in the domain of video understanding. Recent advancements in Video VLMs, such as VideoChat~\citep{li2025videochat} and LLaVA-NeXT-Video~\citep{zhang2024llavanextvideo}, have demonstrated impressive capabilities in comprehending temporally bounded, offline video clips. However, real-world applications often require interacting with continuous, infinite visual inputs on the fly. This necessity has driven a recent paradigm shift towards Streaming Video Large Language Models (Streaming VLMs), which are designed to process newly arriving visual streams in real time and maintain long-term context, as exemplified by StreamingVLM~\citep{ICLR2026_6445dd88}.

\begin{figure}[t]
  \centering
  \includegraphics[width=0.50\textwidth]{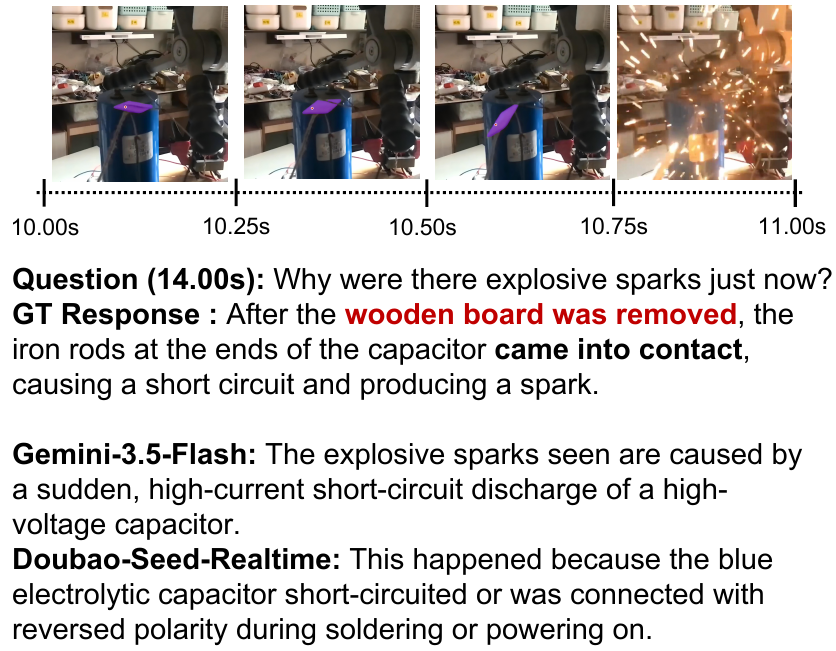}
  \caption{\textbf{A perception failure in a high-dynamic video event.} A wooden separator (highlighted in purple using SAM 3) is removed, allowing the metal rods to touch and produce sparks. Both models provide generic electrical explanations that could be inferred from the sparks alone, but fail to identify the brief sequence of separator removal and subsequent contact that triggers the discharge. This illustrates the challenge of capturing critical visual details in fast-paced events.}
  \label{fig:case}
\end{figure}

Despite this rapid progress, existing streaming evaluation benchmarks, such as OVO-Bench~\citep{Niu_2025_CVPR} and PhoStream~\citep{lu2026phostream}, emphasize temporal reasoning and response timing without explicitly targeting high-dynamic perception. This leaves a critical challenge encountered in real-world streaming applications: accurately perceiving and comprehending highly dynamic, fast-paced events under strict context window constraints. In practice, video streams exhibit a heterogeneous mixture of low- and high-dynamic scenes. Operating under limited token budgets, current models must balance the need for maintaining a long temporal history, preserving high spatial resolution, and capturing fine-grained temporal information. Such constraints motivate sparse frame sampling; for example, the online baselines evaluated in OVO-Bench operate at 1--2 frames per second~\citep{Niu_2025_CVPR}. Consequently, when faced with rapidly changing high-frequency actions—such as sports, extreme gaming, or wildlife hunting—these models often miss crucial visual details, leading to severe perception failures (as illustrated in Fig.~\ref{fig:case}). While recent studies have introduced offline benchmarks focusing on motion comprehension and high frame-per-second (FPS) perception (e.g., MotionBench, \citealp{Hong_2025_CVPR}; FPS-Bench, \citealp{Choudhury_2026_CVPR}), evaluating the capability to perceive high-dynamic events within a continuous, untrimmed streaming context remains an open problem.

To bridge this critical gap, we introduce FastBench, a novel benchmark specifically designed to evaluate the performance of streaming VLMs on high-dynamic real-world video streams. FastBench is constructed via a meticulously designed pipeline that synergizes VLM-driven generation, agentic expert verification, and rigorous human inspection. To ensure true temporal dependency, we introduce a low-FPS verification step that proactively filters out pseudo-dynamic questions. Furthermore, to overcome the inherent hallucination risks of VLMs in fast-paced scenarios, we propose a novel Vision Expert Verification mechanism. By autonomously invoking vision foundation models, the VLM verifies and refines its QA pairs against objective, text-based physical trajectories rather than error-prone visual features. Spanning diverse domains including sports, gaming, nature, and traffic, FastBench establishes a stringent evaluation standard for high-dynamic streaming perception.

Along with the benchmark, we present ProactiveFrame, a simple, training-free baseline framework. Diverging from passive uniform sampling, ProactiveFrame empowers the agent to dynamically adjust its temporal perception granularity via text prompts based on the ongoing streaming context. To balance the need for fine-grained perception and long-term memory under strict token budgets, we introduce a dual-tier context sliding window mechanism. Operating on a single, unified chronological stream, this design proactively buffers high-FPS chunks for fleeting events, which subsequently undergo a graceful temporal degradation into a sparse historical context. This hierarchical management elegantly resolves the trade-off, enabling continuous high-dynamic perception without fragmenting the temporal horizon. Evaluations on FastBench show that high-dynamic perception benefits from denser temporal sampling, but remains constrained by the trade-off between temporal granularity and historical context. ProactiveFrame improves over sparse uniform sampling, yet its gap to oracle-guided sampling highlights the challenge of autonomously identifying when finer temporal perception is needed.

In summary, our main contributions are three-fold:
\begin{itemize}
    \item \textbf{High-Dynamic Streaming Benchmark:} We introduce \textbf{FastBench}, a benchmark specifically designed to evaluate high-frame-rate-dependent perception in continuous video streams under bounded-context and response-timing constraints
    \item \textbf{Trajectory-Grounded Data Pipeline:} We propose an innovative data construction pipeline incorporating a low-FPS filtering mechanism to remove pseudo-dynamic questions, paired with a \textit{Vision Expert Verification} module that grounds VLM annotations on objective physical trajectories to mitigate visual hallucinations.
    \item \textbf{Training-Free Baseline Framework:} We present \textbf{ProactiveFrame}, a simple yet effective training-free framework that features a dual-tier context sliding window mechanism. It dynamically buffers high-FPS chunks for fleeting events and allows them to gracefully degrade into sparse historical context, balancing fine-grained perception and long-term memory under tight token budgets.
\end{itemize}

\begin{figure*}[t]
\centering
\captionsetup[subfigure]{font=small,justification=raggedright,singlelinecheck=false}
\begin{subfigure}[t]{0.59\textwidth}
  \vspace{0pt}
  \centering
  \includegraphics[width=\linewidth,pagebox=cropbox]{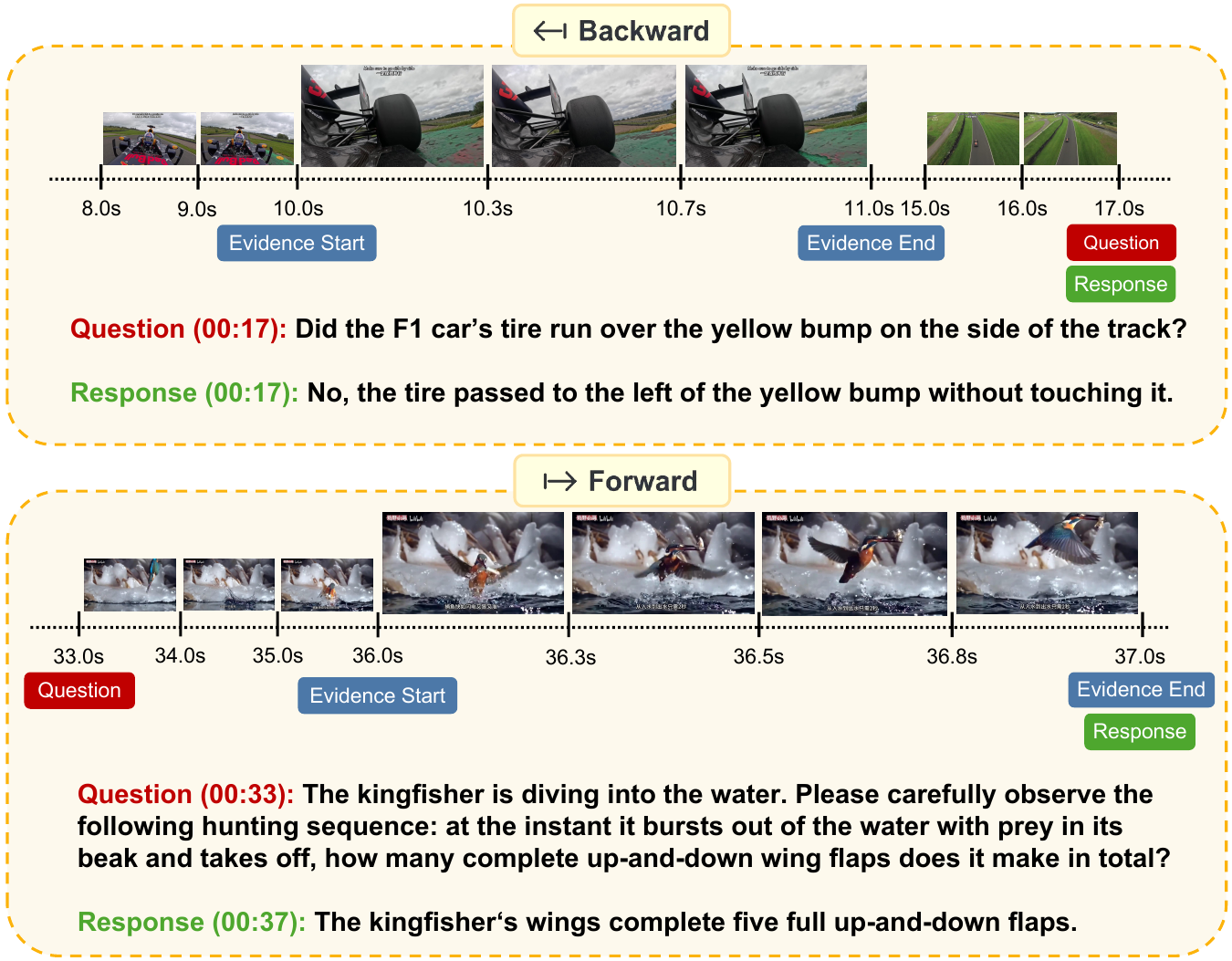}
  \caption{\textbf{Temporal QA examples.} The backward query at 17\,s refers to evidence at 10--11\,s; the forward query at 33\,s is answered after observing evidence at 36--37\,s.}
  \label{fig:dataset_examples}
\end{subfigure}\hfill
\begin{subfigure}[t]{0.39\textwidth}
  \vspace{0pt}
  \centering
  \includegraphics[width=\linewidth]{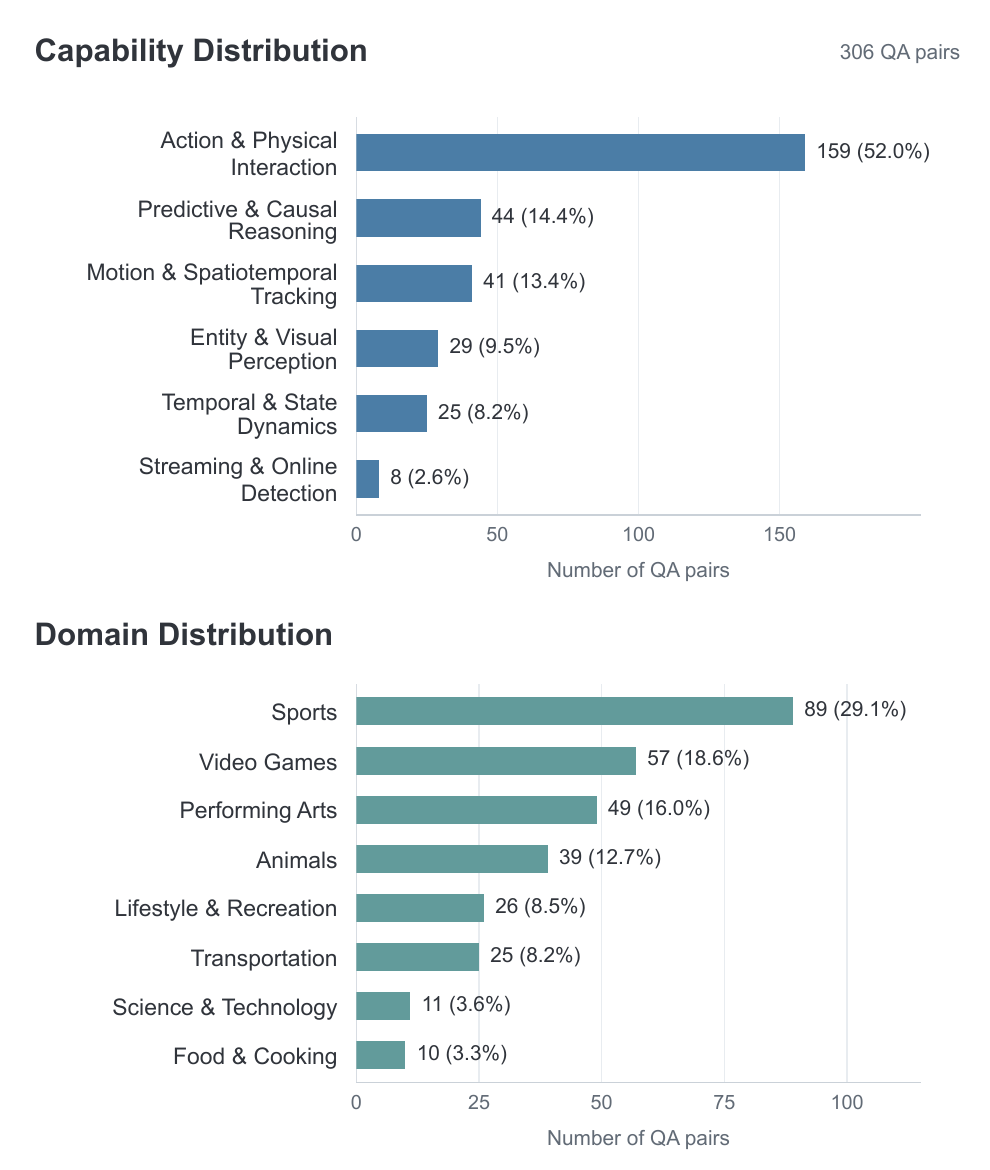}
  \caption{\textbf{Dataset distributions.} Counts and percentages of 306 QA pairs across six capability categories (top) and eight domains (bottom).}
  \label{fig:dataset_distributions}
\end{subfigure}
\caption{\textbf{Overview of FastBench.} Evidence Start and Evidence End in (a) mark the human-annotated interval containing the visual evidence required to answer each question.}
\label{fig:dataset_overview}
\end{figure*}

\section{Related Work}

\subsection{Streaming Video Understanding Benchmarks}

Streaming video benchmarks evaluate understanding from incrementally
available evidence. StreamingBench~\citep{lin2026streamingbench}
assesses real-time visual, omni-source, and contextual understanding,
while OVO-Bench~\citep{Niu_2025_CVPR} covers backward tracing,
real-time understanding, and forward active responding.
OVBench~\citep{huang2025online} evaluates online perception, memory,
and reasoning across past, present, and future contexts.
PhoStream~\citep{lu2026phostream} extends evaluation to open-ended
question answering in mobile scenarios, emphasizing audio-visual
evidence and response timing.
Building on these temporal task structures, FastBench focuses on
high-frame-rate-dependent evidence in continuous streams, jointly
evaluating fine-grained perception, historical retention, and
response timing under a bounded visual context.

\subsection{Fine-Grained Motion and High-Frame-Rate Understanding}

Fine-grained temporal benchmarks seek to distinguish motion
understanding from reliance on static cues.
TempCompass~\citep{liu2024tempcompass} uses videos with shared
static content but contrasting temporal properties.
MotionBench~\citep{Hong_2025_CVPR} evaluates fine-grained motion
comprehension and investigates higher-frame-rate inputs and
motion-preserving feature fusion.
FPS-Bench~\citep{Choudhury_2026_CVPR} introduces minimum
frames-per-second (minFPS), using human annotations to identify
the lowest sampling rate needed to verify an answer.
Complementing these efforts, FastBench uses low-FPS answerability
filtering rather than per-question minFPS estimation to evaluate
high-frame-rate-dependent evidence within a causal streaming protocol.

\subsection{Proactive Perception and Streaming Context Management}

Streaming systems must coordinate response timing, visual
computation, and context retention.
VideoLLM-online~\citep{Chen_2024_CVPR} learns from video streams
to support temporally aligned dialogue.
LION-FS~\citep{li2025lion} separates fast response determination
from detail-enhanced generation, while
StreamMind~\citep{ding2025streammind} uses event-gated LLM invocation
to decouple perception from language reasoning.
VideoChat3~\citep{li2026videochat3} adapts the spatial resolution of
subsequent windows using predicted interaction states.
For context management, StreamingVLM~\citep{ICLR2026_6445dd88}
maintains different KV-cache retention windows for visual and
textual tokens, while simpleStream~\citep{shen2026simple}
demonstrates a strong recent-frame baseline and a
perception--memory trade-off.
In contrast, ProactiveFrame provides a training-free, prompt-driven
baseline that adjusts future temporal sampling rates and uses a
dual-tier window to retain recent high-FPS observations alongside
sparse historical context.

\begin{figure*}[t]
\begin{center}
\includegraphics[width=0.9\textwidth]{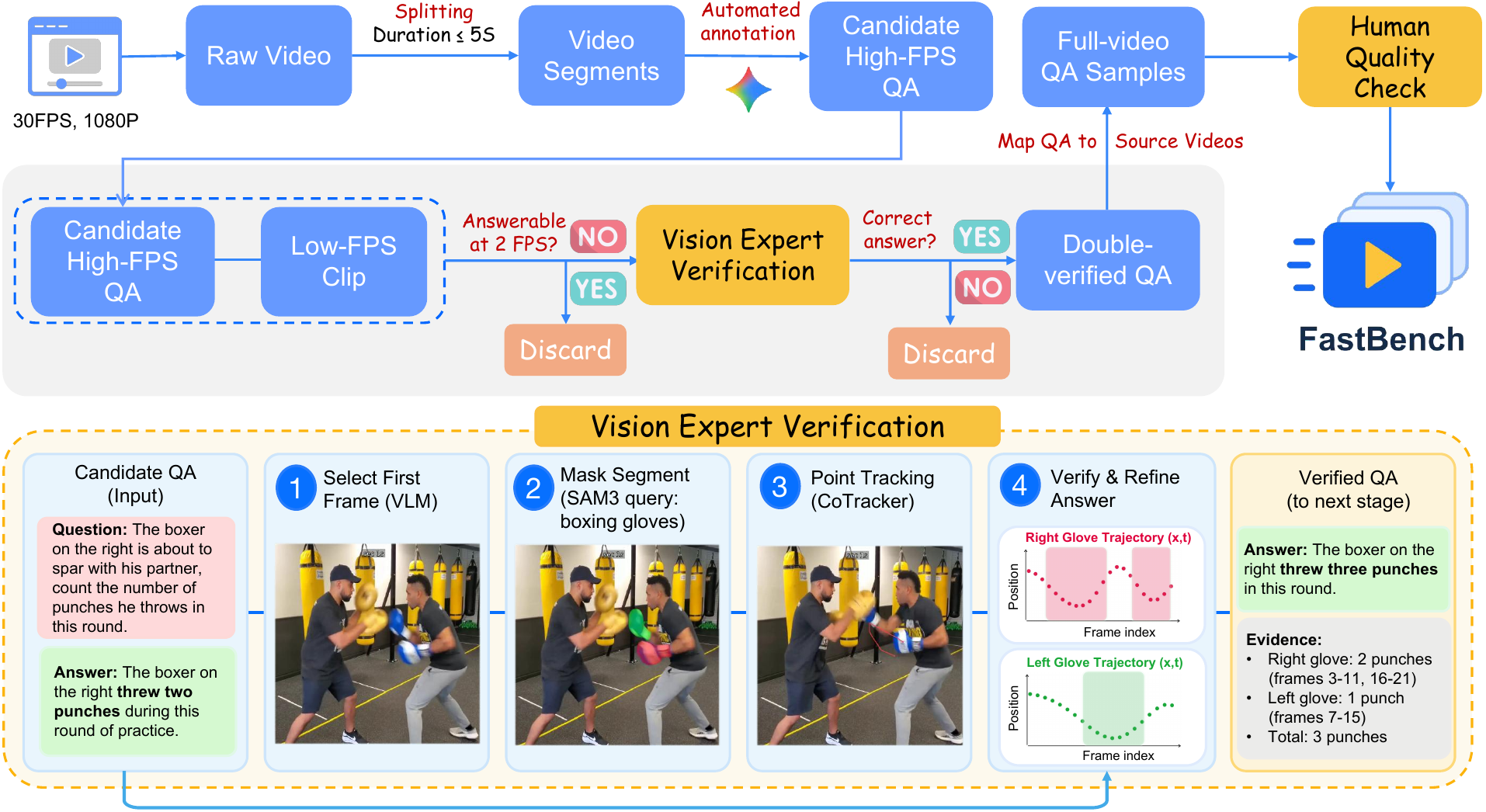} 
\end{center}
\caption{\textbf{FastBench data construction pipeline.} A VLM generates candidate QA pairs from short, high-frame-rate video clips. Questions answerable at 2 FPS are discarded. The remaining pairs undergo automated vision-expert verification, where the VLM uses SAM3 segmentation and CoTracker3 trajectories to verify or refine answers; pairs that cannot be verified or corrected are discarded. Retained QA pairs are mapped back to their source videos and undergo human quality checks to form FastBench. The lower panel illustrates how trajectory evidence supports correcting a punch count from two to three.}
\label{fig:data_pipeline}
\end{figure*}

\section{FastBench}
In this section, we present the design of FastBench and our baseline framework, ProactiveFrame. We first detail the data curation pipeline in Section~\ref{data_curate} and present the dataset statistics in Section~\ref{sec:data_statistic}. Finally, Section~\ref{sec:proactive_frame} describes the framework and core dynamic perception mechanisms of ProactiveFrame.

\subsection{Data Curation}\label{data_curate}
\textbf{Raw Video Collection.} To construct a diverse pool of candidate videos rich in high-dynamic, fast-paced events across domains such as sports, gaming, nature, and traffic, we introduce an automated, metadata-driven collection pipeline. We first employ a VLM, Gemini-3.1-Pro~\citep{google2026gemini31pro}, to brainstorm and generate a comprehensive set of search keywords corresponding to our target high-dynamic scenarios. Using these queries, we retrieve raw, untrimmed videos from popular video-sharing platforms, including YouTube and Bilibili. A metadata filter then uses timestamped comments and danmaku to localize high-dynamic segments within the retrieved videos (Appendix~\ref{app:metadata_filter}).

\textbf{Task Definition.} We formulate FastBench as a Streaming Video Question Answering (QA) task over an untrimmed, continuous video stream $\mathcal{V}$ arriving online. At a query timestamp $t_q$, a user question $Q$ is presented, requiring evidence anchored within a temporal window $[t_{\text{start}}, t_{\text{end}}]$. Following the temporal scopes considered in streaming benchmarks~\citep{Niu_2025_CVPR,lu2026phostream}, we distinguish three cases: \textit{Backward} ($t_{\text{end}} < t_q$), \textit{Instant} ($t_{\text{start}} \le t_q \le t_{\text{end}}$), and \textit{Forward} ($t_{\text{start}} > t_q$). Fig.~\ref{fig:dataset_overview}\subref{fig:dataset_examples} illustrates two of these scopes: the backward example requires recalling whether an F1 tire contacted a track feature before the query, whereas the forward example requires observing and counting a kingfisher's wing flaps after the query.

FastBench explicitly focuses on the unique difficulties of high-dynamic real-world scenarios. Specifically, the benchmark introduces two primary challenges inherent to continuous perception. First, it requires a significantly finer temporal granularity, demanding models to accurately comprehend rapidly evolving visual information across coherent continuous frames rather than isolated snapshots. Second, it highlights a critical trade-off between retaining a long-term historical context and processing high-frequency visual details. Under a limited context token budget, simultaneously accommodating both aspects becomes a severe bottleneck. Given the continuous visual input and query $Q$, the objective of the model is to effectively navigate this context-granularity trade-off to generate an accurate answer $A$.

\textbf{Data Construction Pipeline} We designed a comprehensive data construction pipeline that synergizes VLM-driven generation, agentic expert verification, and rigorous human inspection to extract and annotate high-quality question-answering pairs from high-dynamic video streams, as illustrated in Fig.~\ref{fig:data_pipeline}. Initially, raw high-resolution videos are segmented into short clips with a duration of no more than 5 seconds. For each segment, we maintain a multi-stage offline conversation with a Vision-Language Model to automate the annotation process. In this initial phase, the VLM analyzes the native high-frame-rate video clip and proposes a candidate high-FPS QA pair.

To ensure the generated questions genuinely require continuous perception of high-dynamic events rather than static scene understanding, we introduce a low-FPS verification mechanism. The video clip is temporally sub-sampled at a sparse rate of 2 FPS and fed back into the VLM along with the candidate question. If the model can correctly answer the question relying solely on this low-FPS clip, the question is identified as pseudo-dynamic and is immediately dropped from the pipeline.

Questions that successfully pass the low-FPS filter then enter the Vision Expert Verification stage. The VLM first performs temporal grounding to isolate the target action interval and selects the initial frame of this segment. It then generates a text query for the target object, which is processed by SAM3~\citep{carion2026sam} to produce a segmentation mask. CoTracker3~\citep{Karaev_2025_ICCV} tracks the resulting points over the video segment and outputs their trajectory coordinates. The VLM verifies or corrects the original answer from these trajectories. Unverifiable or uncorrectable QA pairs are discarded, while the successful ones advance as double-verified QA pairs.

\textbf{Human Verification}
Double-verified QA pairs are mapped back to their longer source videos and reviewed by ten experts in multimodal video understanding. Each sample undergoes three rounds of inspection comparing the original high-FPS video with its 2-FPS subsample to verify: (1) answer correctness; (2) clear, unambiguous visual evidence and question targets; and (3) dependence on high-dynamic visual information, ruling out questions answerable at 2 FPS or without visual context. Experts also annotate the evidence interval and specific timestamps requiring high-FPS perception. Fig.~\ref{fig:dataset_overview}\subref{fig:dataset_examples} illustrates these intervals with Evidence Start and Evidence End markers. Only samples passing all three rounds are retained in FastBench.

\begin{figure*}[t]
\begin{center}
\includegraphics[width=0.95\textwidth]{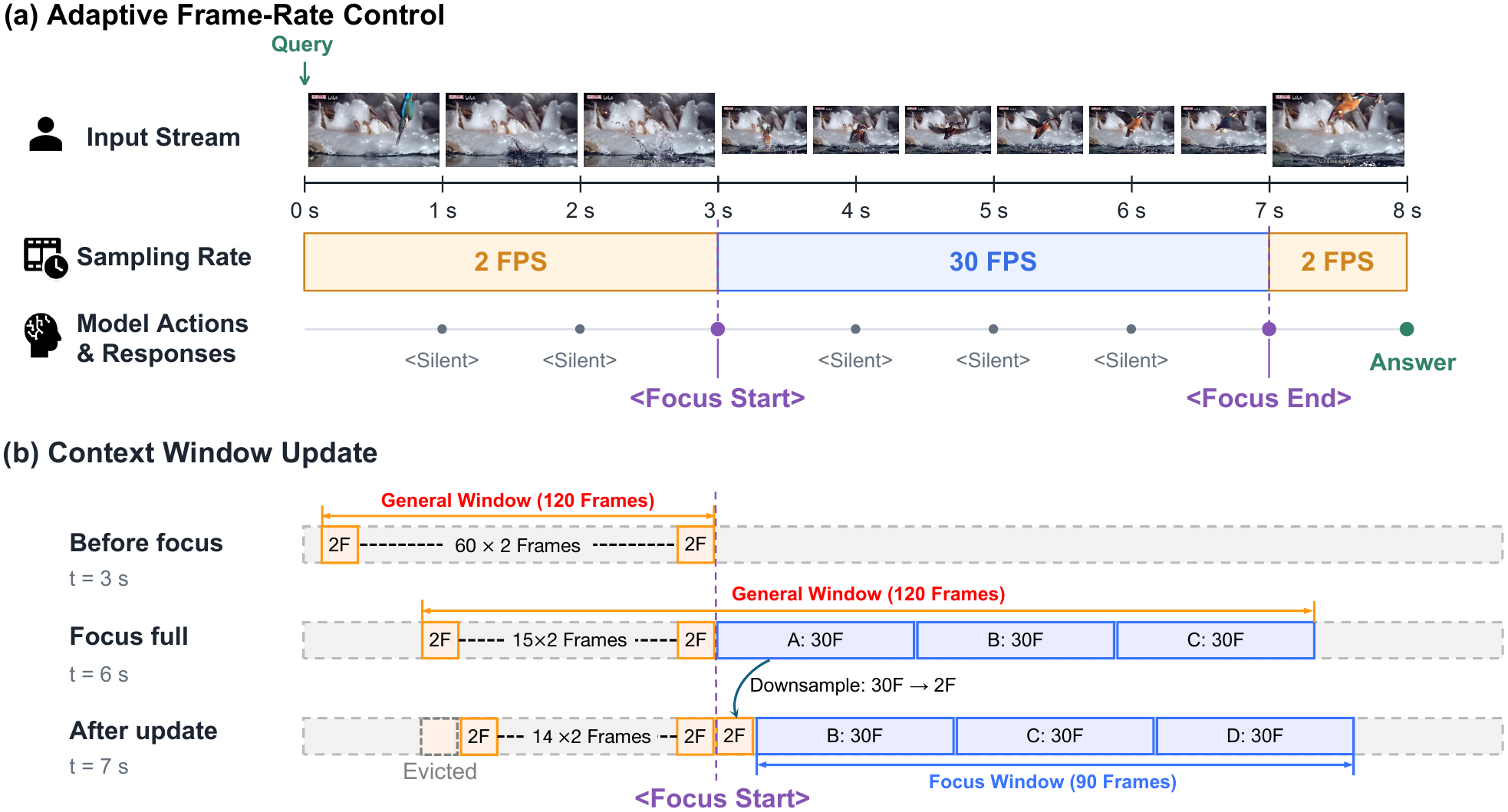} 
\end{center}
\caption{
\textbf{Overview of ProactiveFrame.}
(a) The model switches between low- and high-FPS sampling through
\texttt{<Focus Start>} and \texttt{<Focus End>};
\texttt{<Silent>} preserves the current mode.
(b) A 120-frame General Window includes a 90-frame Focus Window.
When a new high-FPS chunk arrives, the oldest focused chunk is
downsampled into sparse history, and the oldest historical chunk
is evicted to maintain the budget.
Each chunk spans one second; F denotes retained frames.
}
\label{fig:inference_pipeline}
\end{figure*}

\subsection{Data Statistics}\label{sec:data_statistic}

In this section, we present the overall statistics of the FastBench dataset. We first detail the multi-stage data annotation and filtering process. Subsequently, we analyze the dataset's diversity and comprehensively evaluate its distribution across different capability dimensions and real-world domains. The overall statistical distributions for both capabilities and domains are summarized in Fig.~\ref{fig:dataset_overview}\subref{fig:dataset_distributions}.

\textbf{Data Annotation.} To construct FastBench, we employed Gemini-3.1-Pro to generate 89 search keywords for high-dynamic scenarios. For each keyword, a VLM semantically filtered the metadata of the top 200 videos ranked by popularity. This pipeline yielded 766 source videos, segmented into 33,619 clips of up to 5 seconds. The VLM then generated 16,702 candidate question-answering (QA) pairs from these high-frame-rate clips. To ensure temporal dependency and mitigate visual hallucinations, low-FPS filtering and Vision Expert Verification reduced the pool to 3,102 double-verified QAs. Finally, multi-round human inspections for answer correctness, visual clarity, and high-dynamic relevance yielded 306 high-quality samples.

\textbf{Capability Taxonomy and Distribution.} FastBench categorizes its 306 QA pairs into six capability dimensions (Fig.~\ref{fig:dataset_overview}\subref{fig:dataset_distributions}):
(1)~\textit{Action \& Physical Interaction} (51.96\%) covers fine-grained action recognition, physical contact detection, and rapid event counting;
(2)~\textit{Predictive \& Causal Reasoning} (14.38\%) evaluates action anticipation and causal inference;
(3)~\textit{Motion \& Spatiotemporal Tracking} (13.40\%) tests spatial reasoning and trajectory tracking under rapid motion;
(4)~\textit{Entity \& Visual Perception} (9.48\%) assesses object and text recognition under motion blur;
(5)~\textit{Temporal \& State Dynamics} (8.17\%) covers temporal sequencing and sudden state changes; and
(6)~\textit{Streaming \& Online Detection} (2.61\%) targets real-time localization of action boundaries.
Together, these dimensions assess fine-grained perception and higher-level reasoning under streaming constraints.

\textbf{Domain Coverage and Diversity.} FastBench spans eight domains featuring rapid motion and high visual dynamics:
(1)~\textit{Sports} (29.08\%), including motorsports, combat sports, and skateboarding;
(2)~\textit{Video Games} (18.63\%), featuring fast-paced action and esports gameplay;
(3)~\textit{Performing Arts} (16.01\%), including sleight-of-hand magic, card manipulation, and stage music;
(4)~\textit{Animals} (12.75\%), covering swift wildlife movements and hunting;
(5)~\textit{Lifestyle \& Recreation} (8.50\%), encompassing acrobatic stunts, pen spinning, and active hobbies;
(6)~\textit{Transportation} (8.17\%), involving driving, drone operations, and aviation;
(7)~\textit{Science \& Technology} (3.59\%), focusing on high-speed physics experiments; and
(8)~\textit{Food \& Cooking} (3.27\%), covering rapid culinary preparation.
This coverage supports evaluation across varied motion semantics and visual environments.

\subsection{Proactive Frame}\label{sec:proactive_frame}
To resolve the conflict between retaining long-term history and capturing fine-grained temporal details under strict context budgets, we introduce ProactiveFrame, a simple yet effective training-free baseline that enables streaming VLMs to autonomously control their future temporal perception granularity. Instead of passive uniform sampling, the model is guided via text prompts to dynamically switch the frame rate of the incoming stream based on the ongoing streaming context.

Under a fixed context window (e.g., 120 frames), a high frame rate severely compresses the retained history: a 2 FPS stream preserves 60 seconds (sixty 1-second chunks), whereas 30 FPS shrinks the temporal receptive field to merely 4 seconds. To preserve sufficient history while enabling high-dynamic perception, we design a dual-tier context sliding window. As illustrated in Fig.~\ref{fig:inference_pipeline}, assuming a default 1080p resolution, the General Window holds a constant 120 frames, within which a higher-tier Focus Window buffers high-FPS chunks for fleeting events. Its size is adjustable; for example, 90 frames accommodate three consecutive high-FPS (e.g., native 30 FPS) chunks.

When the accumulated high-FPS frames exceed the Focus Window's capacity, the earliest chunk is evicted following a First-In-First-Out (FIFO) principle. Crucially, rather than being discarded, it undergoes graceful temporal degradation: the chunk is down-sampled to a standard low-FPS chunk (e.g., 2 FPS) and enters the lower-tier General Window. This hierarchical management retains a sparse yet continuous long-term history without sacrificing the ability to proactively perceive transient, high-dynamic events.

\begin{table*}[t]
\centering
\caption{Main evaluation results on \textbf{FastBench}. Capabilities are ordered from left to right by sample ratio in descending order. Scores are presented in percentage (\%). Values in green or red parentheses indicate gains or drops relative to their respective base models.}
\label{tab:main_results}
\resizebox{\textwidth}{!}{%
\begin{tabular}{l | cccccc | ccc | c}
\toprule
\multirow{3}{*}{\textbf{Model}} 
& \multicolumn{6}{c|}{\textbf{Capability Breakdown (by Sample Ratio $\downarrow$)}} 
& \multicolumn{3}{c|}{\textbf{Temporal Scope}} 
& \multirow{3}{*}{\textbf{Overall}} \\
\cmidrule(lr){2-7} \cmidrule(lr){8-10}
& \makecell{\textbf{Action \&}\\\textbf{Phys. Inter.}} 
& \makecell{\textbf{Predictive \&}\\\textbf{Causal}} 
& \makecell{\textbf{Motion \&}\\\textbf{Spatiotemp.}} 
& \makecell{\textbf{Entity \&}\\\textbf{Visual}} 
& \makecell{\textbf{Temporal \&}\\\textbf{Dynamics}} 
& \makecell{\textbf{Streaming \&}\\\textbf{Online}} 
& \textbf{Forward} 
& \textbf{Instant} 
& \textbf{Backward} 
& \\
& (51.96\%) & (14.38\%) & (13.40\%) & (9.48\%) & (8.17\%) & (2.61\%) & (54.90\%) & (23.86\%) & (21.24\%) & \\
\midrule
\multicolumn{11}{l}{\textit{\textbf{Closed-source Models}}} \\
\quad \makecell[l]{Doubao-Seed-2.0-Lite\scriptsize\citep{bytedance2026seed20}} & 25.0 & 51.4 & 16.6 & 35.2 & 30.4 & 42.5 & 33.2 & 24.9 & 25.2 & 29.5 \\[2pt]
\quad \makecell[l]{GPT-5.6-Sol\scriptsize\citep{openai2026gpt56sol}} & 42.6 & 56.8 & 57.6 & 45.5 & 43.2 & 30.0 & 45.1 & 51.8 & 44.9 & 46.7 \\[2pt]
\quad \makecell[l]{Gemini-3.5-Flash\scriptsize\citep{google2026gemini35flash}} & 43.6 & 67.3 & 62.9 & 44.8 & 52.0 & 52.5 & 52.7 & 48.5 & 47.7 & 50.7 \\[2pt]
\midrule
\multicolumn{11}{l}{\textit{\textbf{Open-source Models}}} \\
\quad \makecell[l]{Aura (8B)\scriptsize\citep{aura2026}} & 31.1 & 32.3 & 47.8 & 21.4 & 28.8 & 20.0 & 24.9 & 41.4 & 40.3 & 32.1 \\[2pt]
\quad \makecell[l]{Qwen3-VL (8B)\scriptsize\citep{bai2025qwen3vl}} & 30.6 & 34.1 & 47.3 & 32.4 & 32.0 & 2.5 & 33.9 & 34.5 & 28.3 & 32.9 \\[2pt]
\quad \makecell[l]{MOSS-VL-Realtime (11B)\scriptsize\citep{wang2026mossvl}} & 27.3 & 34.5 & 27.8 & 16.6 & 31.2 & 35.0 & 36.1 & 22.2 & 13.2 & 27.9 \\[2pt]
\quad \makecell[l]{JoyAI-VL-Interaction (8B)\scriptsize\citep{yao2026joyai}} & 41.9 & 48.6 & 49.8 & 34.5 & 43.2 & 70.0 & 44.2 & 43.6 & 44.3 & 44.1 \\[2pt]
\midrule
\multicolumn{11}{l}{\textit{\textbf{Proactive Perception Methods}}} \\
\quad \makecell[l]{VideoChat3 (4B)\scriptsize\citep{li2026videochat3}} & 15.7 & 29.5 & 18.5 & 19.3 & 26.4 & 67.5 & 37.6 & 0.0 & 0.0 & 20.7 \\[2pt]
\quad ProactiveFrame (Qwen3-VL 8B) 
& 36.1 \textcolor{ForestGreen}{\scriptsize(+5.5)} 
& 42.7 \textcolor{ForestGreen}{\scriptsize(+8.6)} 
& 47.3 \textcolor{gray}{\scriptsize(0.0)} 
& 33.8 \textcolor{ForestGreen}{\scriptsize(+1.4)} 
& 34.4 \textcolor{ForestGreen}{\scriptsize(+2.4)} 
& 40.0 \textcolor{ForestGreen}{\scriptsize(+37.5)} 
& 43.0 \textcolor{ForestGreen}{\scriptsize(+9.1)} 
& 32.3 \textcolor{red!80!black}{\scriptsize(-2.2)} 
& 32.9 \textcolor{ForestGreen}{\scriptsize(+4.6)} 
& 38.3 \textcolor{ForestGreen}{\scriptsize(+5.4)} \\
\quad ProactiveFrame (Gemini-3.5-Flash) 
& 47.0 \textcolor{ForestGreen}{\scriptsize(+3.4)} 
& 61.4 \textcolor{red!80!black}{\scriptsize(-5.9)} 
& 56.1 \textcolor{red!80!black}{\scriptsize(-6.8)} 
& 56.6 \textcolor{ForestGreen}{\scriptsize(+11.8)} 
& 54.4 \textcolor{ForestGreen}{\scriptsize(+2.4)} 
& 62.5 \textcolor{ForestGreen}{\scriptsize(+10.0)} 
& 53.6 \textcolor{ForestGreen}{\scriptsize(+0.9)} 
& 48.5 \textcolor{gray}{\scriptsize(0.0)} 
& 52.9 \textcolor{ForestGreen}{\scriptsize(+5.2)} 
& 52.2 \textcolor{ForestGreen}{\scriptsize(+1.5)} \\
\bottomrule
\end{tabular}%
}
\end{table*}

\section{Experiment}
\subsection{Experimental Setting}
\textbf{Streaming Evaluation Protocol}. We follow the streaming inference framework introduced by PhoStream~\cite{lu2026phostream}, where videos are streamed to the model in 1-second chunks. At each step, the model must decide whether to continue observing or generate a final response. To achieve this, for general (non-streaming) models, we employ explicit text prompts instructing them to output a "Silent" word if they require more temporal context to answer the query. Conversely, for native streaming models, we preserve their inherent interaction paradigms, evaluating them using their default, model-specific wait tokens to ensure fair and optimal performance. The quality of the final open-ended answers is assessed using an LLM-as-a-Judge.

\textbf{Evaluation Criteria} Early and late responses are assigned a score of 0. However, high-dynamic real-world streams present a unique challenge: for tasks like short-term counting, models typically cannot react at the exact millisecond an event ends. Instead, they require a brief subsequent observation period to confirm the transition into a static state and ensure the action has fully concluded. To accommodate this natural reasoning process, we adapt the strict evaluation criteria by introducing a relaxed late-response boundary, extending the valid temporal window to 4 seconds after the event concludes.

\textbf{Context Window Constraints} We define a unified maximum context budget of 248,832,000 pixels, which is equivalent to the information capacity of 120 frames at a 1080p resolution. For our proposed ProactiveFrame, we configure the focus window to 90 frames (at default 1080p resolution) to buffer high-FPS visual chunks, reserving the remaining capacity for at least 30 low-FPS frames to retain a continuous, sparse historical context.

\subsection{Main Results}
Table~\ref{tab:main_results} presents the main evaluation results on FastBench, revealing a noticeable performance gap between closed-source and open-source models in handling high-dynamic streaming scenarios. Among the evaluated models, Gemini-3.5-Flash achieves the highest overall accuracy within the closed-source category at 50.7\%. In the open-source category, JoyAI-VL-Interaction (8B) leads the performance with an overall score of 44.1\%. Other open-source models demonstrate more modest capabilities on this challenging benchmark, with Qwen3-VL (8B) scoring 32.9\%, Aura achieving 32.1\%, and MOSS-VL-Realtime (11B) yielding an overall score of 27.9\%.

Within the scope of proactive perception methods, VideoChat3 (4B) yields a relatively low overall score of 20.7\%. This underperformance can likely be attributed to the model's primary focus on learning to dynamically control the spatial dimension rather than the temporal dimension. Furthermore, its limited 4B parameter capacity inherently restricts its ability to adequately process and reason over highly dynamic, fast-paced visual events.

In contrast, our proposed zero-shot approach, ProactiveFrame, demonstrates substantial effectiveness. Integrating ProactiveFrame with the Qwen3-VL (8B) baseline significantly improves the model's overall performance to 38.3\%, representing a notable +5.4\% overall increase. This enhancement is driven by massive gains in specific capability breakdowns, highlighted by a remarkable +37.5\% surge in the ``Streaming \& Online'' category and an +8.6\% increase in ``Predictive \& Causal'' reasoning. Furthermore, combining Gemini-3.5-Flash with our proactive perception method surpasses all other evaluated models, achieving a peak overall score of 52.2\%. This configuration also yields impressive sub-category gains, most notably a +11.8\% jump in ``Entity \& Visual'' perception and a +10.0\% boost in ``Streaming \& Online'' tasks.

However, performance drops are also observed in certain sub-categories—for instance, Gemini-3.5-Flash experiences declines in ``Motion \& Spatiotemp.'' (-6.8\%) and ``Predictive \& Causal'' (-5.9\%), while Qwen3-VL exhibits a minor decrease in the ``Instant'' temporal scope (-2.2\%). This suggests that dynamic temporal resampling can occasionally introduce frame-alignment noise or context redundancy during complex spatio-temporal reasoning. Overall, while these results strongly underscore the immense potential of empowering models with autonomous temporal control, they also reveal that current VLMs' proactive perception mechanisms still have substantial room for improvement in precision and stability.

\subsection{Ablation Analysis}
\textbf{Temporal Context Formulation} To examine how streaming models utilize temporal context, Table~\ref{tab:ablation_focus} compares three context formulations: (1) \textit{Full 30 FPS}, processing the stream at its native frame rate; (2) \textit{Mixed FPS (Focus)}, an oracle that provides 30 FPS chunks only within human-annotated focus intervals ($[t_{\text{start}}, t_{\text{end}}]$) and 2 FPS otherwise; and (3) \textit{ProactiveFrame}, where the model triggers focus sampling autonomously from the visual stream. ProactiveFrame improves over the base models in Table~\ref{tab:main_results} (38.3\% vs. 32.9\% for Qwen3-VL (8B) and 52.2\% vs. 50.7\% for Gemini-3.5-Flash), but a gap remains to the Mixed FPS oracle (40.7\% and 57.5\%), indicating that current VLMs still struggle to trigger focus mode at optimal timestamps.

Intriguingly, for Qwen3-VL (8B), \textit{Full 30 FPS} (42.1\%) surpasses the oracle \textit{Mixed FPS} (40.7\%). This reveals a bottleneck in current open-source VLMs: even with perfectly isolated high-FPS focus segments, their fine-grained dynamic understanding remains limited. Consequently, the visual context lost by downsampling background frames outweighs the benefit of longer historical memory.

Temporal scopes further exhibit distinct patterns. On \textit{Forward} queries, Full 30 FPS exceeds Mixed FPS for both models (56.3\% vs. 55.4\% on Gemini-3.5-Flash and 42.0\% vs. 39.4\% on Qwen3-VL). Gemini-3.5-Flash peaks under Full 30 FPS, whereas Qwen3-VL peaks under ProactiveFrame (43.0\%). \textit{Backward} queries peak under the oracle Mixed FPS for both models (55.4\% and 40.6\%). This aligns with task requirements: Forward queries observe events immediately after the query and therefore benefit from dense frames, while Backward queries must retrieve past evidence and are thus sensitive to the memory--resolution trade-off that the oracle balances best.

\textbf{Input Video Frame Rate} To analyze the effect of temporal granularity, we evaluate Qwen3-VL (8B) at input frame rates from 2 to 30 FPS under a unified context budget (Table~\ref{tab:ablation_fps}). Performance scales monotonically from the 2 FPS baseline (32.9\%) up to 24 FPS, peaking at 44.6\% (+11.7\%), confirming that dense sampling is essential for capturing rapid state transitions and fleeting motion cues.

However, native 30 FPS slightly degrades performance to 42.1\%. This saturation reflects a core trade-off in context-constrained streaming VLMs: maximal frame rate severely compresses the historical receptive field and introduces substantial inter-frame redundancy, so 24 FPS offers the best balance between context depth and high-frequency detail.

Capability and scope breakdowns support this observation. Motion-intensive tasks such as \textit{Motion \& Spatiotemporal Tracking} (+13.7\%) and \textit{Predictive \& Causal Reasoning} (+14.5\%) benefit most, peaking at 24 FPS. Meanwhile, the drop at 30 FPS is most pronounced for \textit{Backward} queries (46.2\% at 24 FPS to 37.5\%), since their reliance on historical evidence suffers from the shortened temporal window, whereas \textit{Instant} queries remain strong at 30 FPS (46.3\%), depending mainly on immediate frame resolution.

\begin{table*}[t]
\centering
\caption{Ablation study on temporal context formulation using Qwen3-VL (8B) and Gemini-3.5-Flash on FastBench. Mixed FPS (Focus) provides original 30 FPS video chunks during the annotated focus timestamps and 2 FPS video chunks otherwise.}
\label{tab:ablation_focus}
\resizebox{\textwidth}{!}{%
\begin{tabular}{l | cccccc | ccc | c}
\toprule
\multirow{3}{*}{\textbf{Method}} 
& \multicolumn{6}{c|}{\textbf{Capability Breakdown (by Sample Ratio $\downarrow$)}} 
& \multicolumn{3}{c|}{\textbf{Temporal Scope}} 
& \multirow{3}{*}{\textbf{Overall}} \\
\cmidrule(lr){2-7} \cmidrule(lr){8-10}
& \makecell{\textbf{Action \&}\\\textbf{Phys. Inter.}} 
& \makecell{\textbf{Predictive \&}\\\textbf{Causal}} 
& \makecell{\textbf{Motion \&}\\\textbf{Spatiotemp.}} 
& \makecell{\textbf{Entity \&}\\\textbf{Visual}} 
& \makecell{\textbf{Temporal \&}\\\textbf{Dynamics}} 
& \makecell{\textbf{Streaming \&}\\\textbf{Online}} 
& \textbf{Forward} 
& \textbf{Instant} 
& \textbf{Backward} 
& \\
& (51.96\%) & (14.38\%) & (13.40\%) & (9.48\%) & (8.17\%) & (2.61\%) & (54.90\%) & (23.86\%) & (21.24\%) & \\
\midrule
\multicolumn{11}{c}{\textbf{Qwen3-VL (8B)}} \\
\midrule
Full 30 FPS & 41.0 & 40.9 & 57.6 & 37.2 & 42.4 & 7.5 & 42.0 & 46.3 & 37.5 & 42.1 \\
Proactive Frame & 36.1 & 42.7 & 47.3 & 33.8 & 34.4 & 40.0 & 43.0 & 32.3 & 32.9 & 38.3 \\
Mixed FPS (Focus) & 41.6 & 41.4 & 49.3 & 35.2 & 36.8 & 7.5 & 39.4 & 43.8 & 40.6 & 40.7 \\
\midrule
\multicolumn{11}{c}{\textbf{Gemini-3.5-Flash}} \\
\midrule
Full 30 FPS & 49.4 & 61.4 & 65.9 & 60.0 & 56.8 & 62.5 & 56.3 & 58.1 & 49.5 & 55.3 \\
Proactive Frame & 47.0 & 61.4 & 56.1 & 56.6 & 54.4 & 62.5 & 53.6 & 48.5 & 52.9 & 52.2 \\
Mixed FPS (Focus) & 49.2 & 65.0 & 69.8 & 71.0 & 60.0 & 60.0 & 55.4 & 64.1 & 55.4 & 57.5 \\
\bottomrule
\end{tabular}%
}
\end{table*}

\begin{table*}[t]
\centering
\caption{Ablation study on input video frame rate (FPS) using \textbf{Qwen3-VL (8B)} on \textbf{FastBench}. Capabilities are ordered from left to right by sample ratio in descending order. Scores are presented in percentage (\%). Values in green or red parentheses indicate performance gains or drops relative to the 2 FPS baseline.}
\label{tab:ablation_fps}
\resizebox{\textwidth}{!}{%
\begin{tabular}{l | cccccc | ccc | c}
\toprule
\multirow{3}{*}{\textbf{FPS}} 
& \multicolumn{6}{c|}{\textbf{Capability Breakdown (by Sample Ratio $\downarrow$)}} 
& \multicolumn{3}{c|}{\textbf{Temporal Scope}} 
& \multirow{3}{*}{\textbf{Overall}} \\
\cmidrule(lr){2-7} \cmidrule(lr){8-10}
& \makecell{\textbf{Action \&}\\\textbf{Phys. Inter.}} 
& \makecell{\textbf{Predictive \&}\\\textbf{Causal}} 
& \makecell{\textbf{Motion \&}\\\textbf{Spatiotemp.}} 
& \makecell{\textbf{Entity \&}\\\textbf{Visual}} 
& \makecell{\textbf{Temporal \&}\\\textbf{Dynamics}} 
& \makecell{\textbf{Streaming \&}\\\textbf{Online}} 
& \textbf{Forward} 
& \textbf{Instant} 
& \textbf{Backward} 
& \\
& (51.96\%) & (14.38\%) & (13.40\%) & (9.48\%) & (8.17\%) & (2.61\%) & (54.90\%) & (23.86\%) & (21.24\%) & \\
\midrule
2 FPS & 30.6 & 34.1 & 47.3 & 32.4 & 32.0 & 2.5 & 33.9 & 34.5 & 28.3 & 32.9 \\
4 FPS 
& 37.0 \textcolor{ForestGreen}{\scriptsize(+6.4)} 
& 40.5 \textcolor{ForestGreen}{\scriptsize(+6.4)} 
& 50.2 \textcolor{ForestGreen}{\scriptsize(+2.9)} 
& 35.2 \textcolor{ForestGreen}{\scriptsize(+2.8)} 
& 41.6 \textcolor{ForestGreen}{\scriptsize(+9.6)} 
& 2.5 \textcolor{gray}{\scriptsize(0.0)} 
& 37.4 \textcolor{ForestGreen}{\scriptsize(+3.5)} 
& 41.1 \textcolor{ForestGreen}{\scriptsize(+6.6)} 
& 38.8 \textcolor{ForestGreen}{\scriptsize(+10.5)} 
& 38.6 \textcolor{ForestGreen}{\scriptsize(+5.7)} \\
8 FPS 
& 40.3 \textcolor{ForestGreen}{\scriptsize(+9.7)} 
& 38.6 \textcolor{ForestGreen}{\scriptsize(+4.5)} 
& 54.1 \textcolor{ForestGreen}{\scriptsize(+6.8)} 
& 33.8 \textcolor{ForestGreen}{\scriptsize(+1.4)} 
& 43.2 \textcolor{ForestGreen}{\scriptsize(+11.2)} 
& 0.0 \textcolor{red!80!black}{\scriptsize(-2.5)} 
& 38.1 \textcolor{ForestGreen}{\scriptsize(+4.2)} 
& 44.1 \textcolor{ForestGreen}{\scriptsize(+9.6)} 
& 42.5 \textcolor{ForestGreen}{\scriptsize(+14.2)} 
& 40.5 \textcolor{ForestGreen}{\scriptsize(+7.6)} \\
16 FPS 
& 43.0 \textcolor{ForestGreen}{\scriptsize(+12.4)} 
& 44.5 \textcolor{ForestGreen}{\scriptsize(+10.4)} 
& 59.0 \textcolor{ForestGreen}{\scriptsize(+11.7)} 
& 40.0 \textcolor{ForestGreen}{\scriptsize(+7.6)} 
& 40.0 \textcolor{ForestGreen}{\scriptsize(+8.0)} 
& 2.5 \textcolor{gray}{\scriptsize(0.0)} 
& 41.8 \textcolor{ForestGreen}{\scriptsize(+7.9)} 
& 45.8 \textcolor{ForestGreen}{\scriptsize(+11.3)} 
& 46.8 \textcolor{ForestGreen}{\scriptsize(+18.5)} 
& 43.8 \textcolor{ForestGreen}{\scriptsize(+10.9)} \\
24 FPS 
& 42.9 \textcolor{ForestGreen}{\scriptsize(+12.3)} 
& 48.6 \textcolor{ForestGreen}{\scriptsize(+14.5)} 
& 61.0 \textcolor{ForestGreen}{\scriptsize(+13.7)} 
& 39.3 \textcolor{ForestGreen}{\scriptsize(+6.9)} 
& 41.6 \textcolor{ForestGreen}{\scriptsize(+9.6)} 
& 2.5 \textcolor{gray}{\scriptsize(0.0)} 
& 43.5 \textcolor{ForestGreen}{\scriptsize(+9.6)} 
& 46.0 \textcolor{ForestGreen}{\scriptsize(+11.5)} 
& 46.2 \textcolor{ForestGreen}{\scriptsize(+17.9)} 
& 44.6 \textcolor{ForestGreen}{\scriptsize(+11.7)} \\
30 FPS 
& 41.0 \textcolor{ForestGreen}{\scriptsize(+10.4)} 
& 40.9 \textcolor{ForestGreen}{\scriptsize(+6.8)} 
& 57.6 \textcolor{ForestGreen}{\scriptsize(+10.3)} 
& 37.2 \textcolor{ForestGreen}{\scriptsize(+4.8)} 
& 42.4 \textcolor{ForestGreen}{\scriptsize(+10.4)} 
& 7.5 \textcolor{ForestGreen}{\scriptsize(+5.0)} 
& 42.0 \textcolor{ForestGreen}{\scriptsize(+8.1)} 
& 46.3 \textcolor{ForestGreen}{\scriptsize(+11.8)} 
& 37.5 \textcolor{ForestGreen}{\scriptsize(+9.2)} 
& 42.1 \textcolor{ForestGreen}{\scriptsize(+9.2)} \\
\bottomrule
\end{tabular}%
}
\end{table*}

\section{Conclusion}

We introduced FastBench, a benchmark for evaluating streaming VLMs on high-dynamic real-world streams, built with a trajectory-grounded pipeline that filters pseudo-dynamic questions and verifies answers against SAM3 and CoTracker3 trajectories before multi-round human inspection. We also presented ProactiveFrame, a training-free baseline that adjusts future frame rates via text tokens and retains history through a dual-tier sliding window. Experiments show that high-dynamic perception remains largely unsolved: the best model reaches only 50.7\%, denser sampling helps but is bounded by the granularity--history trade-off, and ProactiveFrame improves over sparse sampling yet trails oracle-guided focusing, indicating that deciding \emph{when} to look closer is the central open challenge. We hope FastBench and its annotated focus timestamps foster future research on proactive temporal control in streaming video understanding.

\subsection*{AI Use Statement}

In this work, we used Gemini-3.1-Pro~\citep{google2026gemini31pro} to construct the FastBench question-answer annotations in Figure~\ref{fig:data_pipeline}. It proposed the search keywords used to collect high-dynamic source videos, and then annotated each short high-frame-rate clip: it proposed a candidate question-answer pair, discarded the pair when the question was answerable from a 2 FPS subsample, and otherwise selected the action interval, queried SAM3 for a segmentation mask, and verified or revised the answer against CoTracker3 trajectories. Pairs that could not be verified or corrected were discarded. Gemini-3.5-Flash~\citep{google2026gemini35flash} was used to read timestamped comments and danmaku when localizing candidate segments. Every pair retained in the benchmark was afterwards accepted or rejected by human experts. We also used an LLM-as-a-judge to score open-ended answers, following the same protocol as PhoStream~\citep{lu2026phostream}. We have not used generative AI tools to develop theoretical models, formulate or prove mathematical claims, propose hypotheses, design the method, or translate text. We reviewed the model-generated annotations against the source videos and the trajectory evidence, and human experts made the final decision for every released sample. We take responsibility for the final content of this work, including the annotations and evaluation judgments produced with the aid of generative AI.

\clearpage
\bibliography{iclr2027_conference}
\bibliographystyle{iclr2027_conference}

\clearpage
\appendix
\section{Appendix}
\setcounter{table}{0}
\renewcommand{\thetable}{\thesection.\arabic{table}}
\setcounter{figure}{0}
\renewcommand{\thefigure}{\thesection.\arabic{figure}}
\subsection{More Statistics of FastBench}
\label{app:statistics}
\begin{figure}[htbp]
\centering
\includegraphics[width=\textwidth]{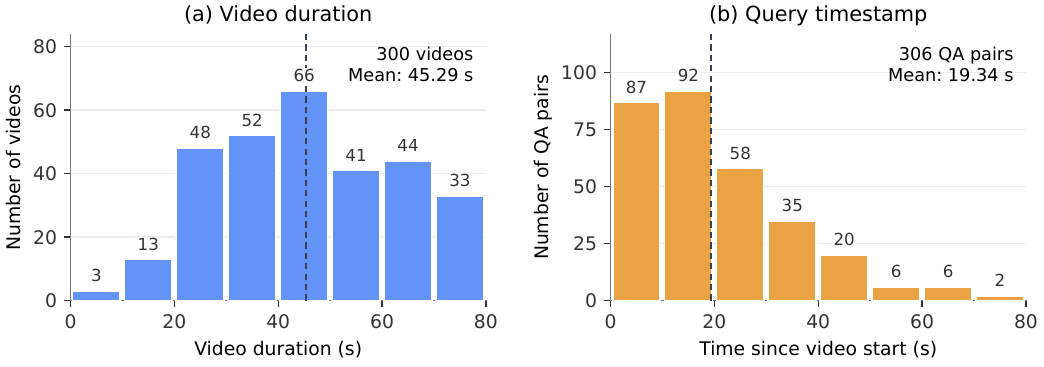}
\caption{\textbf{Additional temporal statistics of FastBench.} (a) Duration distribution of the 300 video clips. (b) Query timestamp distribution of the 306 QA pairs, measured from the start of each clip. Both panels use 10-second bins; bar labels indicate counts and dashed lines mark the means.}
\label{fig:appendix_statistics}
\end{figure}

FastBench contains 306 QA pairs over 300 video clips: 294 clips contain one QA pair each, and six contain two. The clips have a cumulative duration of 3.77 hours, with a mean of 45.29 seconds and a range of 7.32--79.68 seconds. Figure~\ref{fig:appendix_statistics} summarizes the distributions of clip durations and query timestamps. Queries are issued 19.34 seconds after the start of a clip on average, with timestamps ranging from 0 to 74 seconds. These statistics complement the capability and domain distributions in Section~\ref{sec:data_statistic} by characterizing the temporal context in which questions are presented.

\subsection{Details of Human Annotators}
\label{app:annotators}

The human verification in Section~\ref{data_curate} was conducted by ten annotators with experience in multimodal video understanding, each holding at least a master's degree. Annotators were compensated at \$20 per hour.

\subsection{More Detailed Evaluation Results}
\label{app:detailed_results}
We provide additional evaluation results across domains and capability--temporal scope combinations, using the same protocol and scoring scale as Table~\ref{tab:main_results}.

\textbf{Results across domains.} Table~\ref{tab:appendix_domains} reports scores across the eight domains in FastBench, ordered by sample count. Each domain score is averaged over its constituent QA pairs, while Overall is averaged over all 306 QA pairs.

\begin{table}[htbp]
\centering
\caption{\textbf{Evaluation results across domains.} Scores use a 0--100 scale. Parentheses in the header indicate the number of QA pairs in each domain. Best displayed scores in each column are bolded.}
\label{tab:appendix_domains}
\small
\setlength{\tabcolsep}{3pt}
\renewcommand{\arraystretch}{1.12}
\resizebox{\textwidth}{!}{%
\begin{tabular}{l|cccccccc|c}
\toprule
\textbf{Model} & \makecell{\textbf{Sports}\\(89)} & \makecell{\textbf{Video}\\\textbf{Games}\\(57)} & \makecell{\textbf{Performing}\\\textbf{Arts}\\(49)} & \makecell{\textbf{Animals}\\(39)} & \makecell{\textbf{Lifestyle \&}\\\textbf{Recreation}\\(26)} & \makecell{\textbf{Transpor-}\\\textbf{tation}\\(25)} & \makecell{\textbf{Science \&}\\\textbf{Technology}\\(11)} & \makecell{\textbf{Food \&}\\\textbf{Cooking}\\(10)} & \textbf{Overall} \\
\midrule
\multicolumn{10}{l}{\textit{\textbf{Closed-source Models}}} \\
Doubao-Seed-2.0-Lite & 38.4 & 28.1 & 20.4 & 28.2 & 15.4 & 41.6 & 23.6 & 22.0 & 29.5 \\
GPT-5.6-Sol & 53.7 & 37.5 & 44.9 & 45.1 & 36.2 & 48.0 & \textbf{65.5} & \textbf{54.0} & 46.7 \\
Gemini-3.5-Flash & 56.4 & 43.2 & \textbf{49.8} & \textbf{52.3} & \textbf{46.2} & 49.6 & \textbf{65.5} & 38.0 & 50.7 \\
\midrule
\multicolumn{10}{l}{\textit{\textbf{Open-source Models}}} \\
Aura (8B) & 36.0 & 29.1 & 29.0 & 27.7 & 40.0 & 28.0 & 41.8 & 26.0 & 32.1 \\
Qwen3-VL (8B) & 38.4 & 35.4 & 26.1 & 27.7 & 26.9 & 25.6 & 50.9 & 36.0 & 32.9 \\
MOSS-VL-Realtime (11B) & 28.5 & 25.3 & 29.8 & 31.3 & 24.6 & 23.2 & 27.3 & 36.0 & 27.9 \\
\midrule
\multicolumn{10}{l}{\textit{\textbf{Proactive Perception Methods}}} \\
VideoChat3 (4B) & 21.8 & 21.1 & 13.9 & 30.3 & 13.1 & 21.6 & 27.3 & 14.0 & 20.7 \\
\makecell[l]{ProactiveFrame\\(Qwen3-VL 8B)} & 40.4 & 35.8 & 37.1 & 47.7 & 36.2 & 29.6 & 41.8 & 26.0 & 38.3 \\
\makecell[l]{ProactiveFrame\\(Gemini-3.5-Flash)} & \textbf{62.5} & \textbf{49.5} & 47.8 & 48.7 & 43.1 & \textbf{50.4} & \textbf{65.5} & 26.0 & \textbf{52.2} \\
\bottomrule
\end{tabular}%
}
\end{table}

\textbf{Capability-wise results by temporal scope.} Table~\ref{tab:appendix_capability_scope} further separates each capability into Forward, Instant, and Backward subsets, complementing the aggregate capability and temporal scores in Table~\ref{tab:main_results}. All eight Streaming \& Online Detection samples belong to Forward; its Instant and Backward subsets are empty.

\begin{table}[p]
\centering
\caption{\textbf{Capability-wise evaluation results by temporal scope.} Scores use a 0--100 scale. Fwd., Inst., and Back. denote Forward, Instant, and Backward, respectively. A dash indicates an empty subset. Best displayed scores in each column are bolded.}
\label{tab:appendix_capability_scope}
\small
\setlength{\tabcolsep}{4pt}
\renewcommand{\arraystretch}{1.12}
\resizebox{\textwidth}{!}{%
\begin{tabular}{l|ccc|ccc|ccc}
\toprule
\textbf{Model} & \multicolumn{3}{c|}{\makecell{\textbf{Action \&}\\\textbf{Physical Interaction}}} & \multicolumn{3}{c|}{\makecell{\textbf{Predictive \&}\\\textbf{Causal Reasoning}}} & \multicolumn{3}{c}{\makecell{\textbf{Motion \&}\\\textbf{Spatiotemporal Tracking}}} \\
\cmidrule(lr){2-4}\cmidrule(lr){5-7}\cmidrule(lr){8-10}
 & \textbf{Fwd.} & \textbf{Inst.} & \textbf{Back.} & \textbf{Fwd.} & \textbf{Inst.} & \textbf{Back.} & \textbf{Fwd.} & \textbf{Inst.} & \textbf{Back.} \\
\midrule
\multicolumn{10}{l}{\textit{\textbf{Closed-source Models}}} \\
Doubao-Seed-2.0-Lite & 26.0 & 29.5 & 19.6 & 51.7 & \textbf{40.0} & 56.0 & 12.6 & 16.0 & 28.6 \\
GPT-5.6-Sol & 41.1 & \textbf{49.5} & 39.1 & 55.6 & \textbf{40.0} & \textbf{76.0} & 50.5 & \textbf{65.3} & 60.0 \\
Gemini-3.5-Flash & 43.8 & 48.0 & 39.6 & \textbf{69.4} & 26.7 & \textbf{76.0} & \textbf{60.0} & 60.0 & 77.1 \\
\midrule
\multicolumn{10}{l}{\textit{\textbf{Open-source Models}}} \\
Aura (8B) & 23.0 & 41.0 & 35.2 & 28.9 & 20.0 & 64.0 & 40.0 & 48.0 & 68.6 \\
Qwen3-VL (8B) & 33.4 & 29.5 & 27.0 & 36.7 & 20.0 & 24.0 & 44.2 & 46.7 & 57.1 \\
MOSS-VL-Realtime (11B) & 37.0 & 22.0 & 16.5 & 36.7 & \textbf{40.0} & 16.0 & 34.7 & 30.7 & 2.9 \\
\midrule
\multicolumn{10}{l}{\textit{\textbf{Proactive Perception Methods}}} \\
VideoChat3 (4B) & 34.2 & 0.0 & 0.0 & 36.1 & 0.0 & 0.0 & 40.0 & 0.0 & 0.0 \\
\makecell[l]{ProactiveFrame\\(Qwen3-VL 8B)} & 44.9 & 28.5 & 28.7 & 45.0 & \textbf{40.0} & 28.0 & 46.3 & 37.3 & 71.4 \\
\makecell[l]{ProactiveFrame\\(Gemini-3.5-Flash)} & \textbf{47.7} & 47.5 & \textbf{45.7} & 61.1 & \textbf{40.0} & \textbf{76.0} & 49.5 & 50.7 & \textbf{85.7} \\
\bottomrule
\end{tabular}%
}
\par\vspace{1em}
\resizebox{\textwidth}{!}{%
\begin{tabular}{l|ccc|ccc|ccc}
\toprule
\textbf{Model} & \multicolumn{3}{c|}{\makecell{\textbf{Entity \&}\\\textbf{Visual Perception}}} & \multicolumn{3}{c|}{\makecell{\textbf{Temporal \&}\\\textbf{State Dynamics}}} & \multicolumn{3}{c}{\makecell{\textbf{Streaming \&}\\\textbf{Online Detection}}} \\
\cmidrule(lr){2-4}\cmidrule(lr){5-7}\cmidrule(lr){8-10}
 & \textbf{Fwd.} & \textbf{Inst.} & \textbf{Back.} & \textbf{Fwd.} & \textbf{Inst.} & \textbf{Back.} & \textbf{Fwd.} & \textbf{Inst.} & \textbf{Back.} \\
\midrule
\multicolumn{10}{l}{\textit{\textbf{Closed-source Models}}} \\
Doubao-Seed-2.0-Lite & 38.9 & 17.5 & 60.0 & 38.6 & 20.0 & 20.0 & 42.5 & \textemdash & \textemdash \\
GPT-5.6-Sol & 41.1 & \textbf{50.0} & 60.0 & 45.7 & 42.9 & 35.0 & 30.0 & \textemdash & \textemdash \\
Gemini-3.5-Flash & 51.1 & 25.0 & 60.0 & 48.6 & \textbf{62.9} & \textbf{45.0} & 52.5 & \textemdash & \textemdash \\
\midrule
\multicolumn{10}{l}{\textit{\textbf{Open-source Models}}} \\
Aura (8B) & 13.3 & 40.0 & 20.0 & 21.4 & 40.0 & 35.0 & 20.0 & \textemdash & \textemdash \\
Qwen3-VL (8B) & 30.0 & 45.0 & 13.3 & 38.6 & 31.4 & 10.0 & 2.5 & \textemdash & \textemdash \\
MOSS-VL-Realtime (11B) & 24.4 & 5.0 & 0.0 & 47.1 & 17.1 & 0.0 & 35.0 & \textemdash & \textemdash \\
\midrule
\multicolumn{10}{l}{\textit{\textbf{Proactive Perception Methods}}} \\
VideoChat3 (4B) & 31.1 & 0.0 & 0.0 & 47.1 & 0.0 & 0.0 & \textbf{67.5} & \textemdash & \textemdash \\
\makecell[l]{ProactiveFrame\\(Qwen3-VL 8B)} & 33.3 & 40.0 & 20.0 & 37.1 & 31.4 & 30.0 & 40.0 & \textemdash & \textemdash \\
\makecell[l]{ProactiveFrame\\(Gemini-3.5-Flash)} & \textbf{60.0} & 42.5 & \textbf{73.3} & \textbf{57.1} & 60.0 & 35.0 & 62.5 & \textemdash & \textemdash \\
\bottomrule
\end{tabular}%
}
\end{table}

\subsection{Qualitative Dataset Examples}
\label{app:qualitative_examples}

\begin{figure*}[p]
\centering
\captionsetup[subfigure]{font=small,justification=raggedright,singlelinecheck=false}

\begin{subfigure}{\textwidth}
\centering
\includegraphics[height=0.18\textheight]{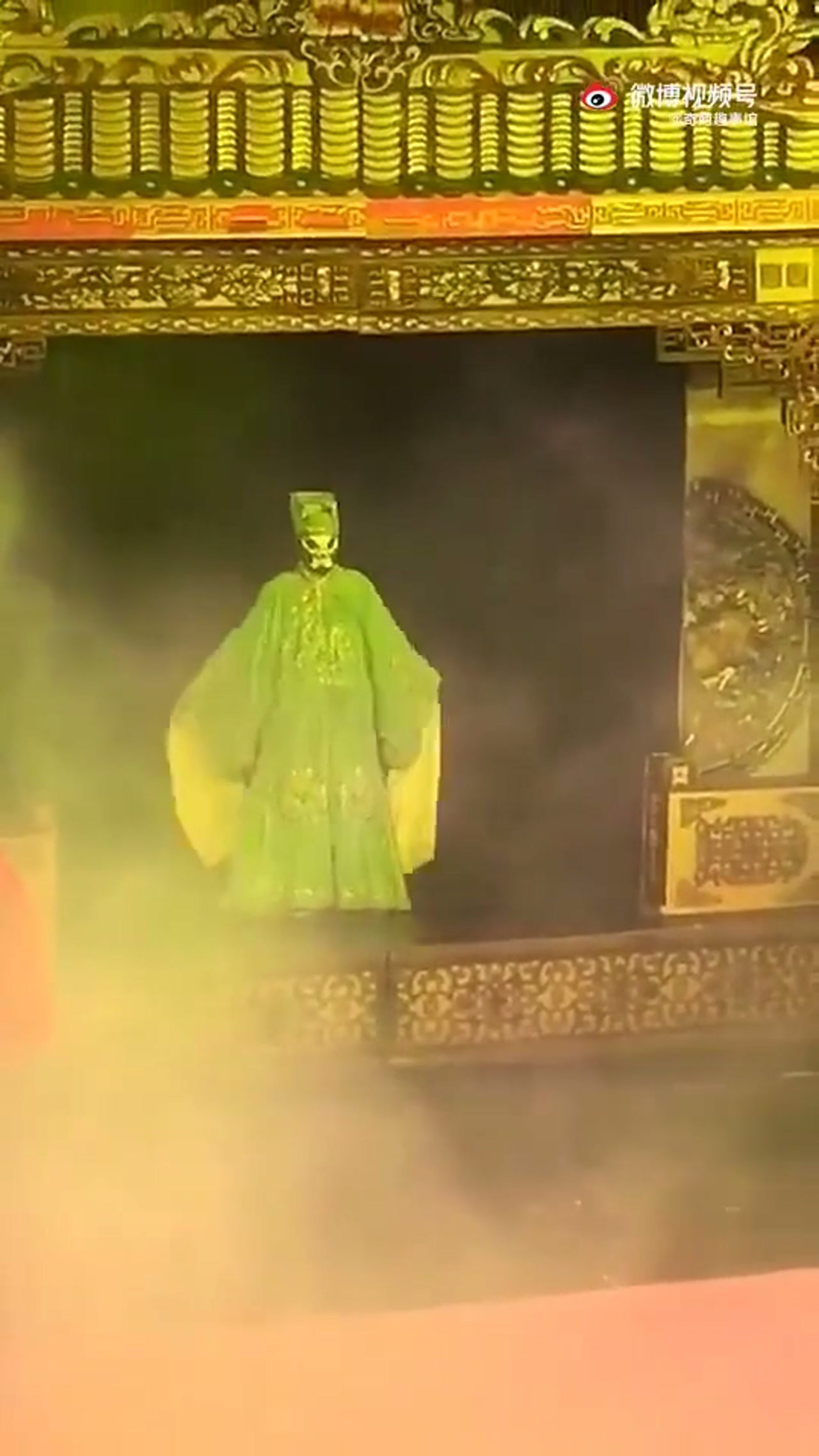}\hfill
\includegraphics[height=0.18\textheight]{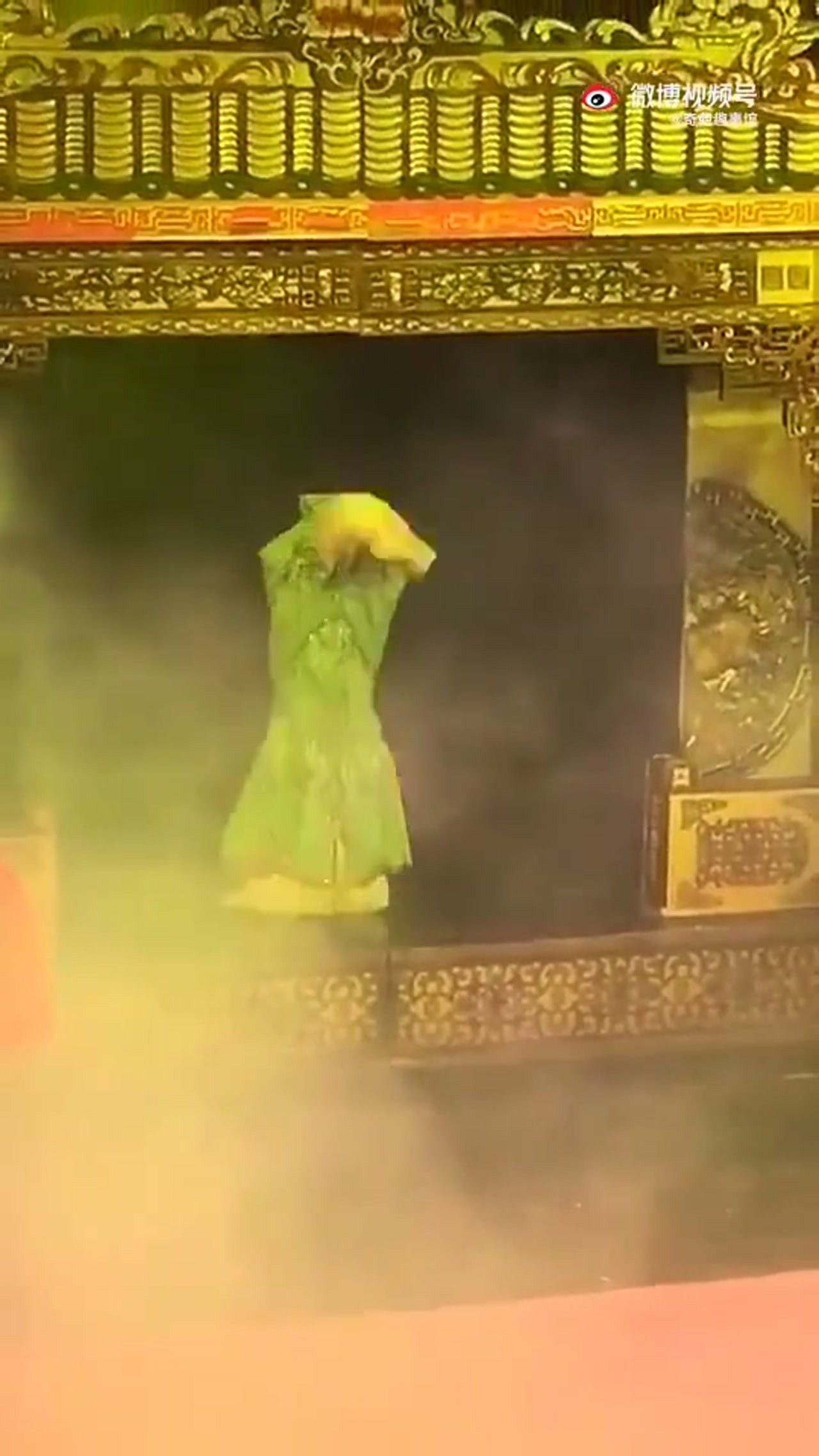}\hfill
\includegraphics[height=0.18\textheight]{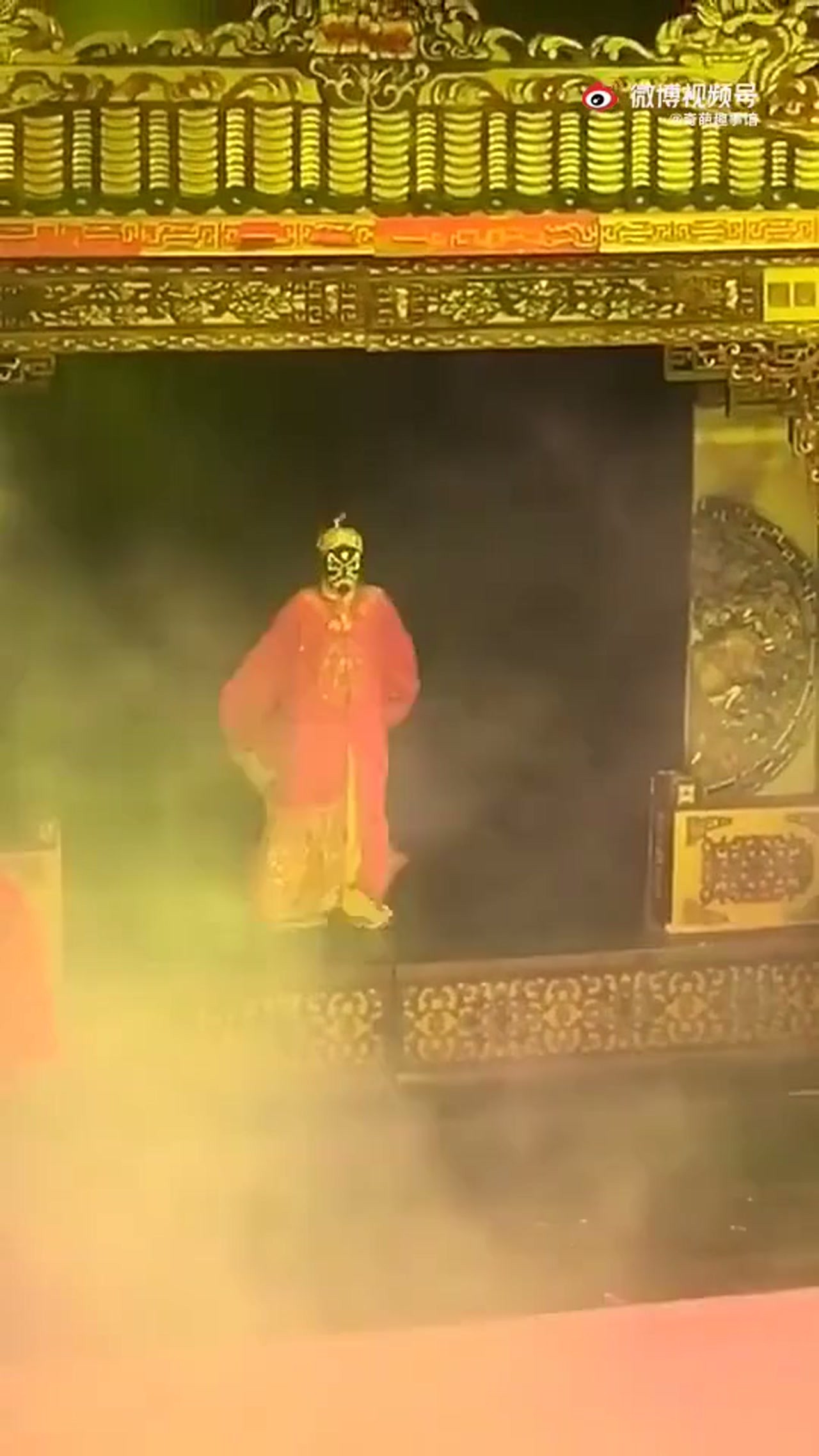}
\caption{\textbf{Performing arts, forward} (sample-58). Query and focus at 3\,s; the answer is available at 5\,s. Frames at 4.0\,s, 4.4\,s, and 5.0\,s. \textit{Q:} Look closely at this actor, what surprising move will he make next? \textit{A:} He raises both arms, covers his face with his sleeves, and as the arms drop his mask and costume change from green to red.}
\label{fig:appendix_case_face}
\end{subfigure}

\vspace{0.6em}
\begin{subfigure}{\textwidth}
\centering
\includegraphics[height=0.18\textheight]{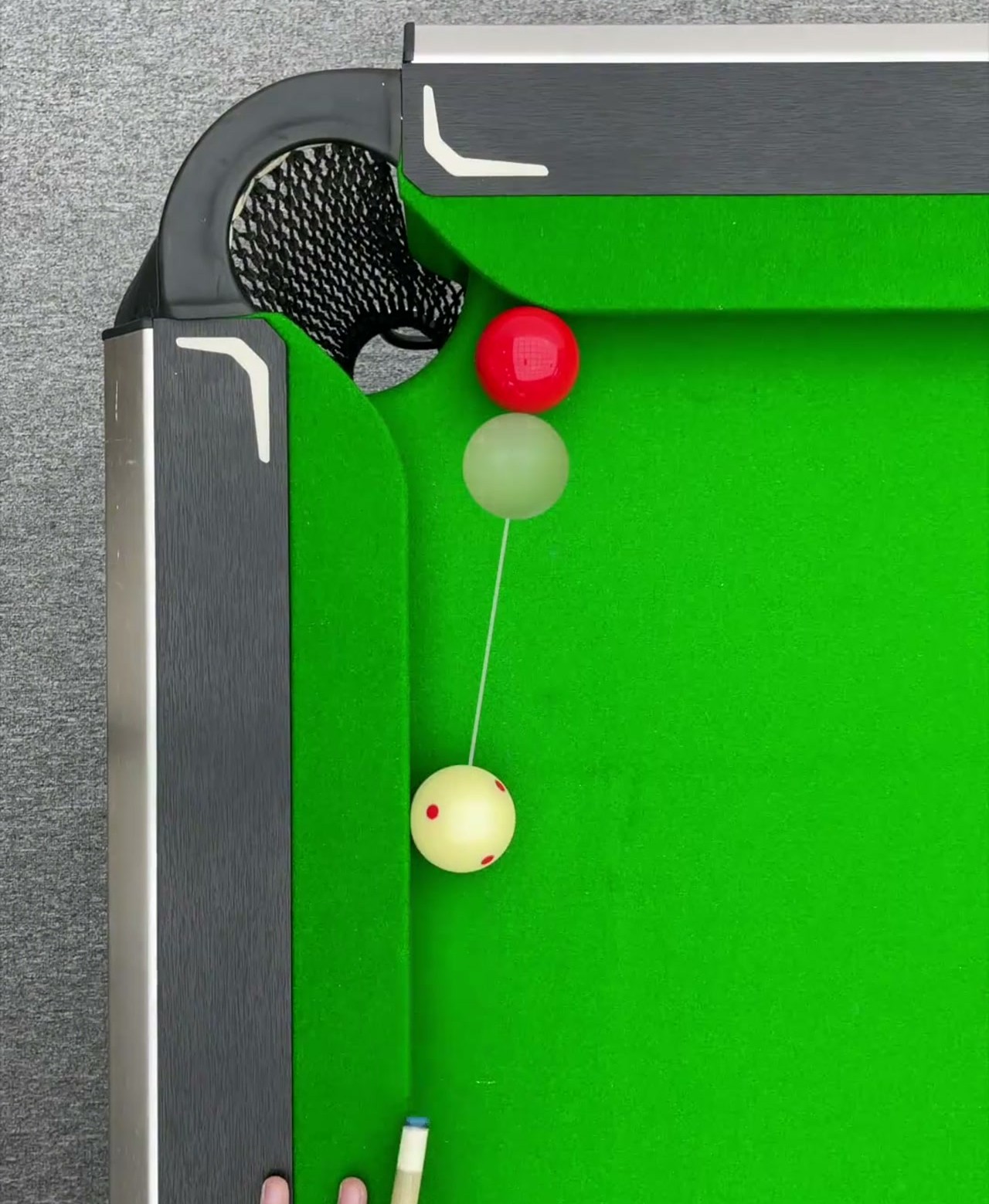}\hfill
\includegraphics[height=0.18\textheight]{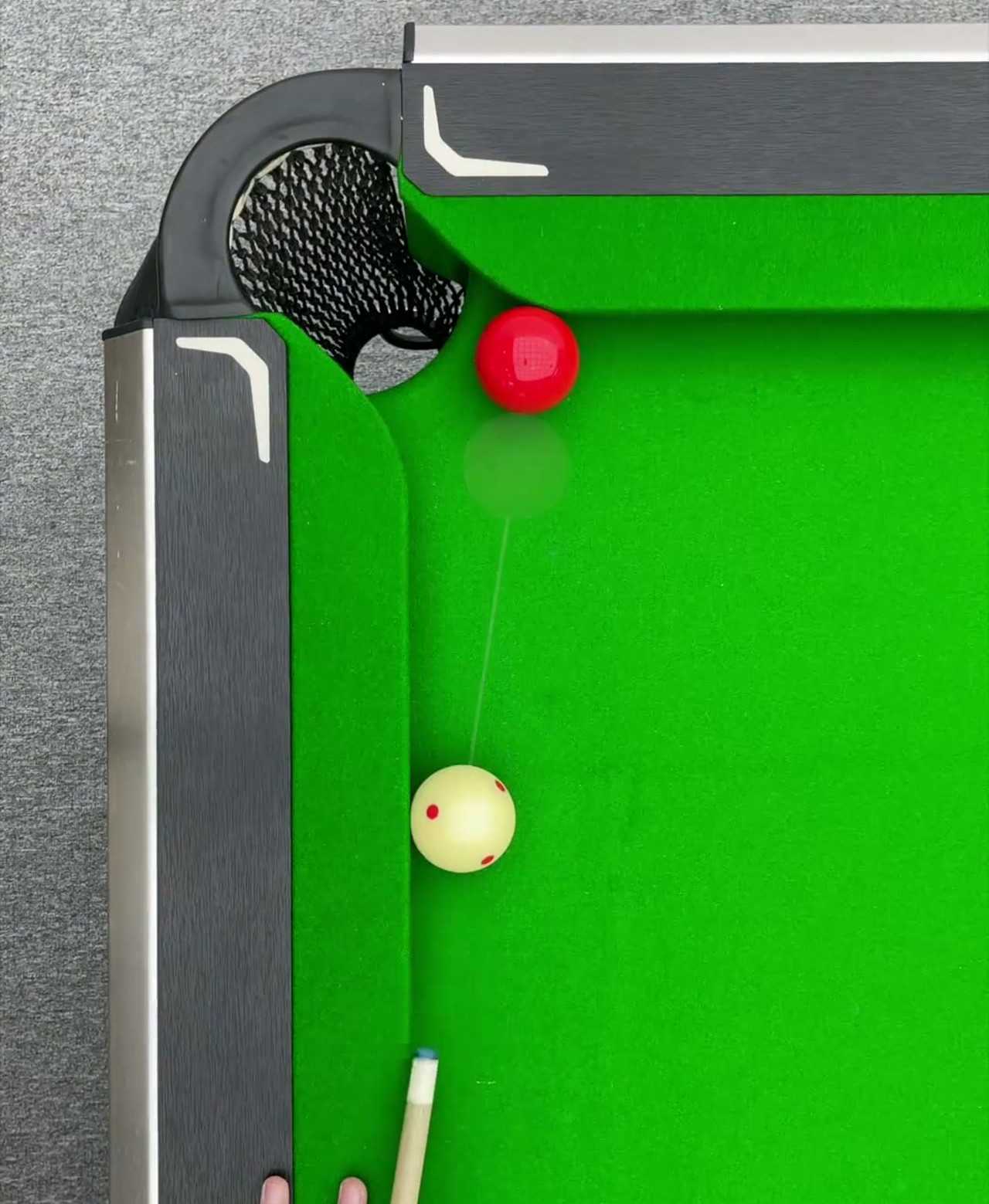}\hfill
\includegraphics[height=0.18\textheight]{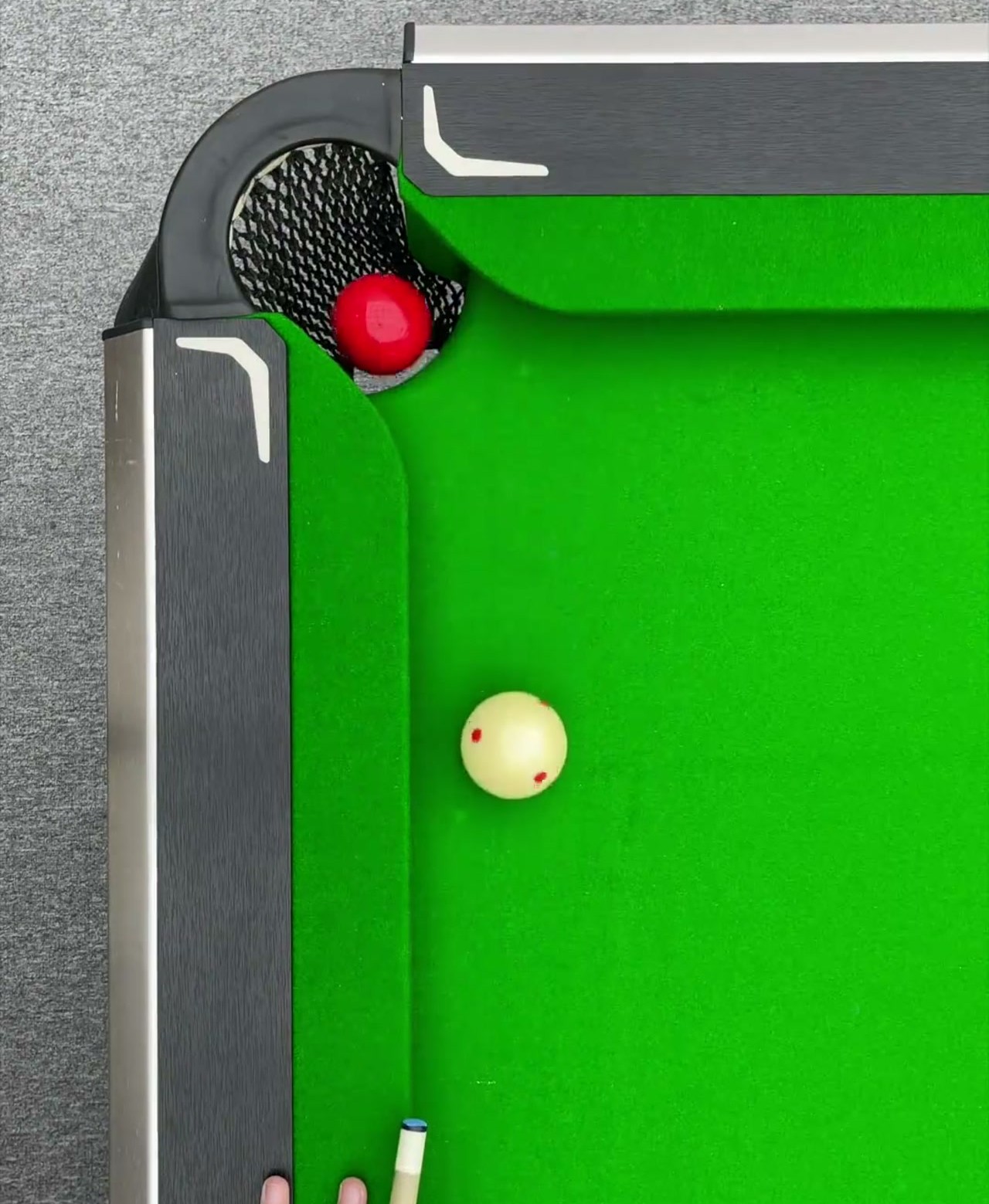}
\caption{\textbf{Lifestyle and recreation, backward} (sample-14). Focus at 16\,s; query at 17\,s. Frames at 16.0\,s, 16.3\,s, and 17.3\,s. \textit{Q:} The white ball is rolling toward the red ball. Will the red ball be knocked into the top-left pocket? \textit{A:} Yes. The white ball strikes the red ball, which rolls left and drops into the top-left pocket.}
\label{fig:appendix_case_billiards}
\end{subfigure}

\vspace{0.6em}
\begin{subfigure}{\textwidth}
\centering
\includegraphics[width=0.32\textwidth]{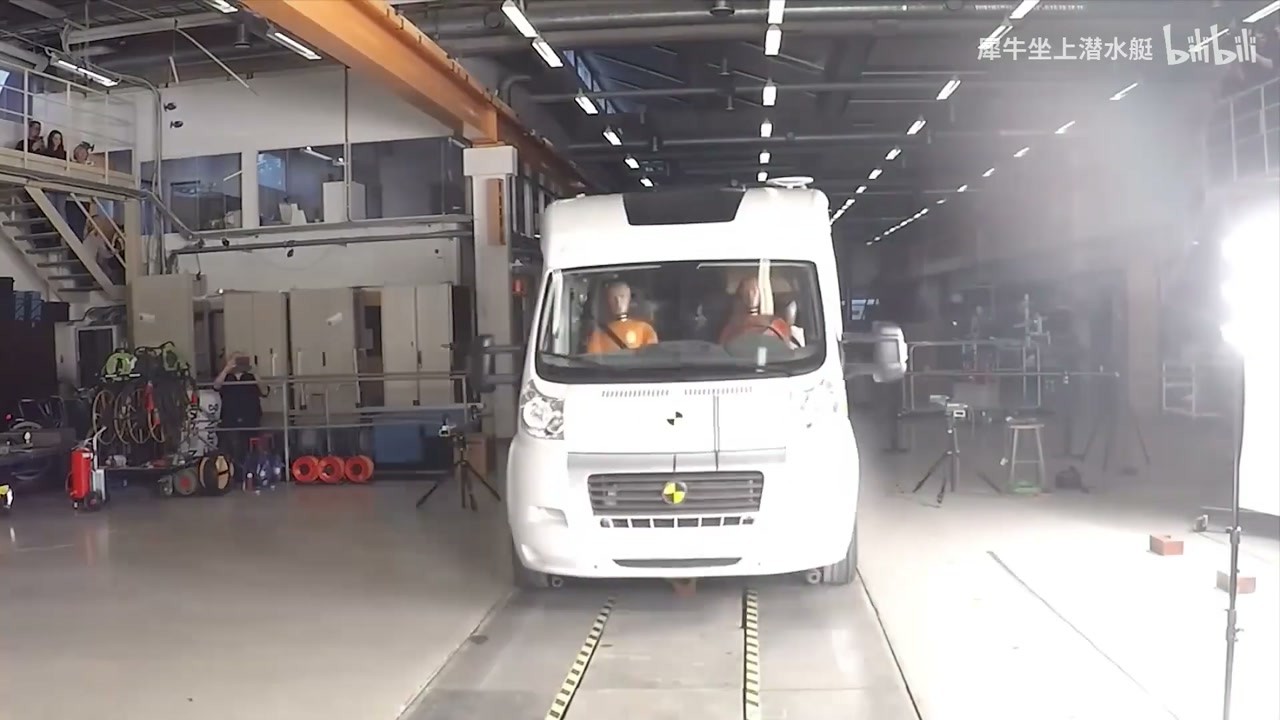}\hfill
\includegraphics[width=0.32\textwidth]{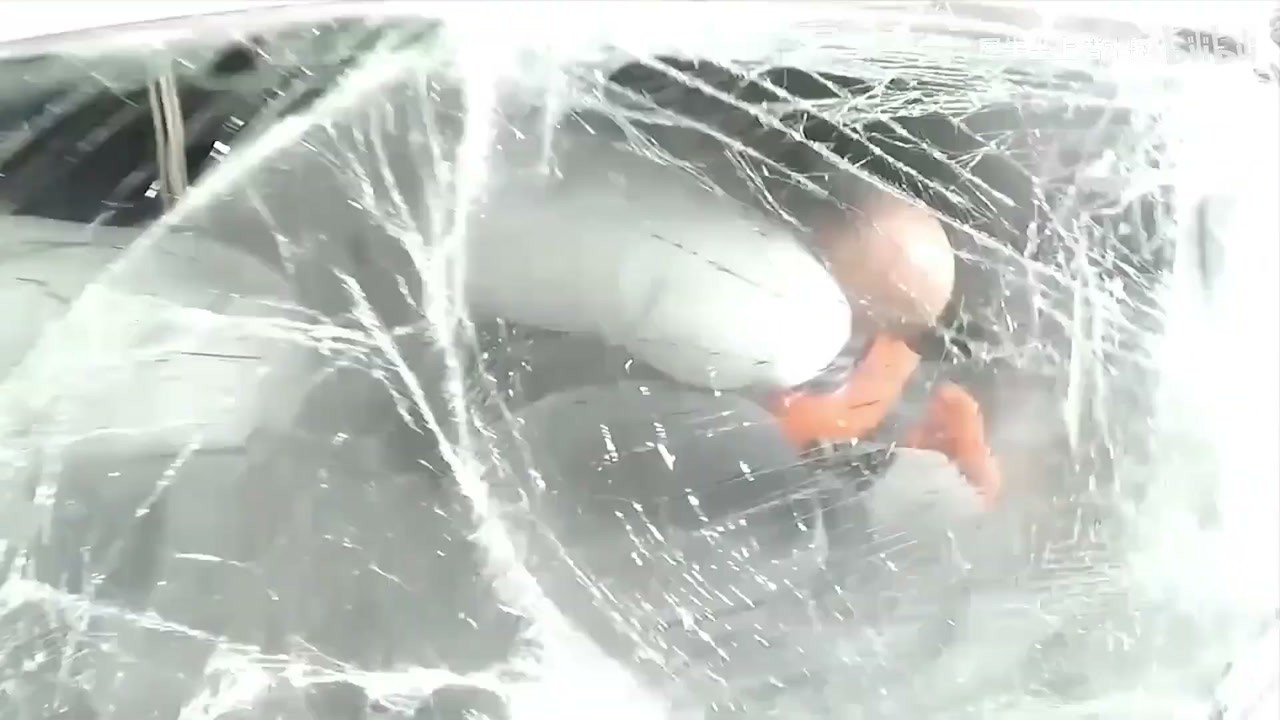}\hfill
\includegraphics[width=0.32\textwidth]{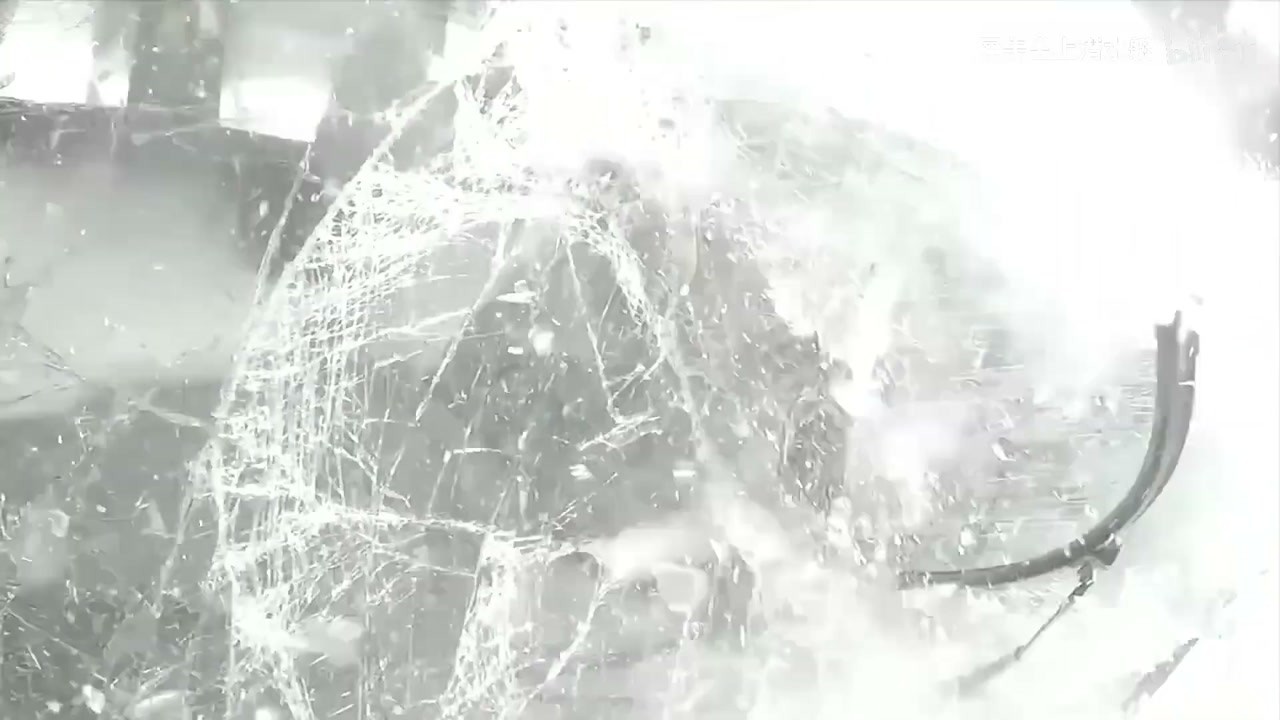}
\caption{\textbf{Transportation, backward} (sample-212). Focus at 7\,s; query at 8\,s. Frames at 7.0\,s, 7.4\,s, and 7.8\,s. \textit{Q:} At the frontal collision, what happens to the black windshield wiper, and which way does it fly off? \textit{A:} It jolts loose, rotates, and flies off toward the right side of the screen.}
\label{fig:appendix_case_wiper}
\end{subfigure}

\vspace{0.6em}
\begin{subfigure}{\textwidth}
\centering
\includegraphics[width=0.32\textwidth]{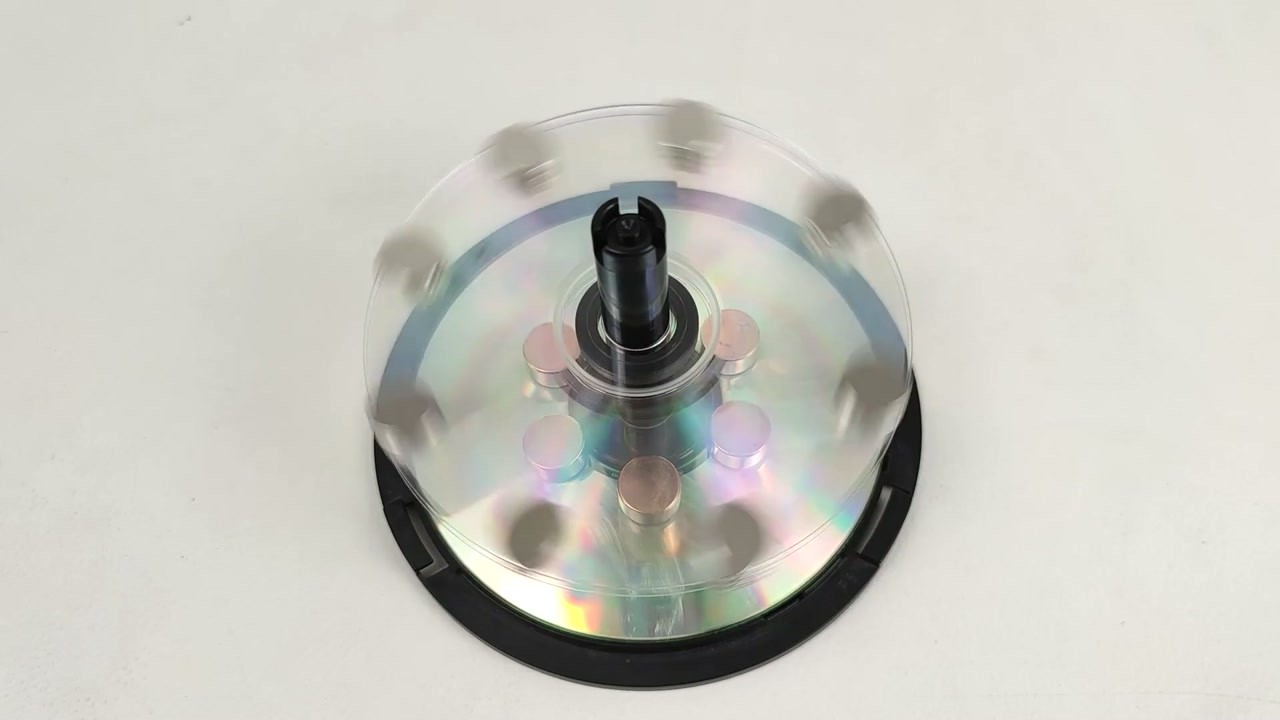}\hfill
\includegraphics[width=0.32\textwidth]{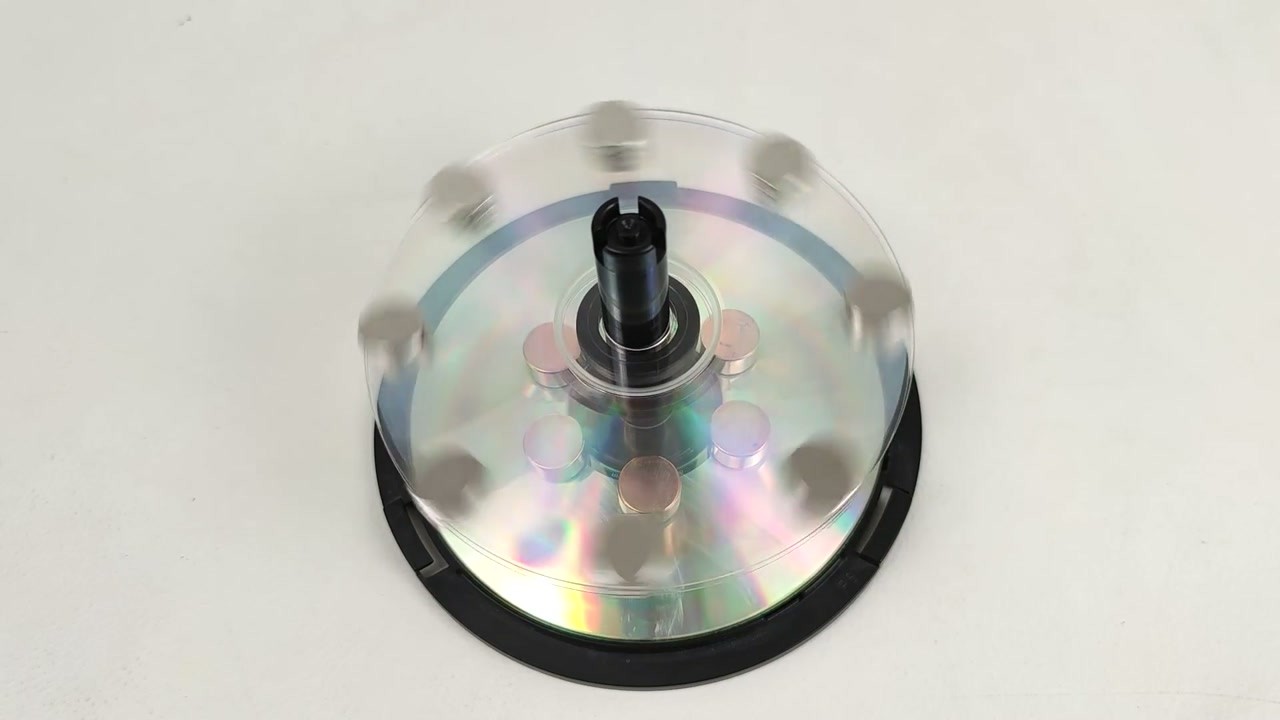}\hfill
\includegraphics[width=0.32\textwidth]{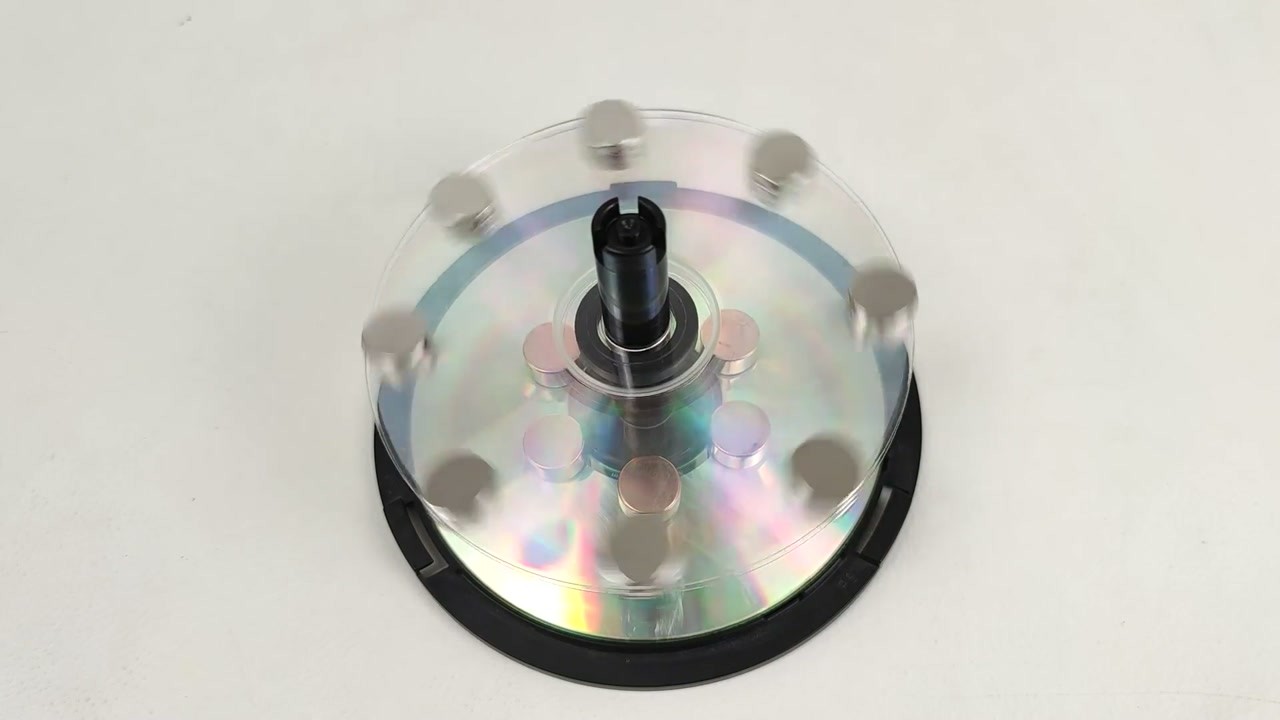}
\caption{\textbf{Science and technology, backward} (sample-31). Focus at 64\,s; query at 67\,s. Frames at 64\,s, 65\,s, and 67\,s. \textit{Q:} Was the disc just rotating clockwise or counterclockwise? \textit{A:} Clockwise. The spinning frames are motion-blurred and the last frame is already still, so the direction is not recoverable from these three images.}
\label{fig:appendix_case_disc}
\end{subfigure}

\caption{\textbf{Qualitative examples from FastBench.} Each row is one QA pair. Focus is the annotated timestamp that requires high-frame-rate perception; the query time is when the question is asked.}
\label{fig:appendix_examples}
\end{figure*}

Figure~\ref{fig:appendix_examples} shows four FastBench items whose answers depend on a brief visual change. Each row contains three frames from the evidence interval, with the query time, focus timestamp, and annotated answer.

\subsection{Metadata Filtering}
\label{app:metadata_filter}

This subsection details the metadata filter that follows keyword retrieval in Section~\ref{data_curate}.

Identifying fleeting, high-dynamic moments within hours of raw video using pure visual processing is computationally prohibitive. However, we empirically observe a strong correlation between specific user engagement patterns and the occurrence of rapid, visually complex events. Notably, timestamped comments (e.g., ``01:02 What just flashed by?'') and time-aligned bullet chats frequently highlight moments that capture intense audience attention, serving as natural indicators of high-dynamic scenes. Leveraging this insight, we formulate a VLM-assisted filtering mechanism. We feed the retrieved video metadata—specifically the comments and danmaku—into Gemini-3.5-Flash~\citep{google2026gemini35flash}. The VLM acts as an efficient semantic filter, analyzing these textual cues to not only determine the presence of high-dynamic content but also to accurately pinpoint the temporal coordinates of these target segments. This step efficiently prunes the video corpus and extracts precise temporal windows for the subsequent verification stages.

Figure~\ref{fig:metadata_filter_prompt} gives the instruction used for this filter. The model must return a JSON match decision and, when the video matches, an approximate timestamp for the window passed to later verification.

\begin{figure}[htbp]
\centering
\begin{minipage}{\textwidth}
\footnotesize
\begin{verbatim}
# Task
Analyze the given video metadata (title, sampled danmaku, and
comments) and determine whether the video contains a high-dynamic
transient: a sudden switch from an ordinary, redundant state to a
moment that requires high-FPS, frame-by-frame understanding (for
example, an ultra-fast serve, bursting a balloon, an extreme dodge,
or an exposed magic trick). If so, precisely localize the
approximate time span in which this high-dynamic transient occurs.

# Criteria (target scene characteristics)
1. Abruptness: the video is mostly calm or shows only routine
   motion, but at one point an extremely fast and complex action
   happens suddenly.
2. Low-FPS ambiguity: at a conventional low frame rate the moment
   is severely motion-blurred and can be resolved only frame by
   frame or in slow motion.

# Signal Indicators (cues from multiple sources)
- Title cues: words such as "extreme", "clutch", "sudden",
  "split-second capture", "instant", or "teleport".
- Comment and danmaku behavior cues: mentions of "0.25x speed",
  "pause", "rewind", "couldn't see it", "in a blink", or
  "couldn't screenshot it".
- Comment timestamps: explicit markers such as "01:15 watch this"
  or "the last 3 seconds".
- Danmaku timing cues: a dense burst of "???", "so fast", or
  "what just happened".
Filter the video according to the requirements above, and provide
both the reasoning and the target segment time. The output must be
valid JSON and contain exactly three fields:
{"is_match": <true|false>,
 "reasoning": "<at most 120 characters>",
 "target_timestamp": "<MM:SS / HH:MM:SS / None>"}.
If the time cannot be determined, output None. Do not output any
additional text.
\end{verbatim}
\end{minipage}
\caption{Instruction for the metadata filter used before visual inspection. It reads the title, sampled danmaku, and comments, decides whether the video contains a sudden high-dynamic moment that requires frame-level perception, and returns an approximate timestamp of that moment.}
\label{fig:metadata_filter_prompt}
\end{figure}

\subsection{ProactiveFrame Instruction}
\label{app:proactiveframe_prompt}

Figure~\ref{fig:proactiveframe_prompt} shows the instruction used by ProactiveFrame during streaming inference (Section~\ref{sec:proactive_frame}). On each new video chunk, the model answers the active question when the observed context is sufficient; otherwise it emits \texttt{Focus\_Start} or \texttt{Focus\_End} as the entire response to raise or restore the frame rate of the next chunk, or outputs \texttt{Silent} to keep the current mode.

\begin{figure}[p]
\centering
\begin{minipage}{\textwidth}
\footnotesize
\setlength{\baselineskip}{9.2pt}
\begin{verbatim}
You are an AI assistant for real-time video stream question
answering. At any moment, there is at most one active question.

Answer the active question as soon as the current video
context—based on all frames observed so far and prior
dialogue—contains enough information to support a complete and
factually grounded answer.

Stream input semantics:
- New visual input is appended incrementally in chunks.
- Each newly appended chunk covers at most 1 second of new
  video time.
- The number of frames in a newly appended chunk can vary; do
  not assume a fixed frame count per update.
- You can proactively request a temporary higher-frame-rate
  view when the current sampling rate may miss fine-grained
  temporal details.

Follow these output rules exactly:

- Focus control:
  1) Proactively monitor the video even when there is no active
     question. Output the exact token Focus_Start whenever the
     current scene appears likely to contain an upcoming
     highlight, rapid motion, a brief event, a transition, an
     interaction, object manipulation, a subtle gesture, or any
     detail that could benefit from fine-grained temporal
     perception. This preserves richer evidence for possible
     future backward-looking or instant questions.
  2) Also output Focus_Start when an active question requires
     closer temporal observation before it can be answered
     reliably. High-frame-rate sampling starts with the next
     video chunk.
  3) Use a low threshold for starting focus. When uncertain
     whether higher frame rate may help, prefer Focus_Start;
     unnecessary focus is acceptable.
  4) While high-frame-rate sampling is active, inspect each new
     chunk and output the exact token Focus_End once the
     interesting/high-motion/fine-grained interval has clearly
     ended and the scene has settled. Normal sampling resumes
     with the next video chunk. Until then, keep focus active
     by answering the active question if possible, or outputting
     Silent.
  5) Focus_Start and Focus_End must each be the entire response.
     Do not combine either token with an answer, Silent,
     punctuation, or explanation.
  6) Do not output Focus_Start if high-frame-rate sampling is
     already active. Do not output Focus_End unless it is
     active.

- For multiple-choice questions (those that explicitly list
  options such as A, B, C, D):
  Respond only with the correct option letter (e.g., "A"). Do
  not add punctuation, explanation, or extra text.

- For all other questions:
  Provide a complete answer using only information available in
  the stream up to this point.
  If the answer can be inferred with reasonable confidence,
  answer directly.

- Output the exact word Silent only in one of these cases:
  1) There is no active question and neither Focus_Start nor
     Focus_End is appropriate.
  2) User explicitly asks to delay answering.
  3) Question unambiguously requires future content and no part
     can be determined yet.
\end{verbatim}
\end{minipage}
\caption{Instruction used by ProactiveFrame to control frame rate from the stream itself. \texttt{Focus\_Start} requests a higher frame rate for the next chunk, \texttt{Focus\_End} returns to normal sampling, and each must be the entire response; \texttt{Silent} continues the current mode when the model should not yet answer.}
\label{fig:proactiveframe_prompt}
\end{figure}

\end{document}